\pdfoutput=1
\documentclass[10pt,twocolumn,letterpaper]{article}

\usepackage[pagenumbers]{cvpr} % To force page numbers, e.g. for an arXiv version

\usepackage[accsupp]{axessibility} % Improves PDF readability for those with disabilities.
\usepackage{caption}

\usepackage{wrapfig}
\usepackage{makecell} 
\usepackage[dvipsnames]{xcolor}

\usepackage{multirow} 
\usepackage{floatrow}
\newfloatcommand{capbtabbox}{table}[][\FBwidth]
\usepackage{soul} 
\usepackage{color, xcolor} 

\definecolor{cvprblue}{rgb}{0.21,0.49,0.74}
\usepackage[pagebackref,breaklinks,colorlinks,citecolor=cvprblue]{hyperref}

\def\paperID{947} % *** Enter the Paper ID here
\def\confName{CVPR}
\def\confYear{2024}

\title{A Unified Framework for Microscopy Defocus Deblur with Multi-Pyramid Transformer and Contrastive Learning}  % \footnotemark[1]

\author{
Yuelin Zhang$^1$ \and
Pengyu Zheng$^1$ \and
Wanquan Yan$^1$ \and
Chengyu Fang$^2$ \and
Shing Shin Cheng$^1$\footnotemark[2] \and \\
$^1$ Department of Mechanical and Automation Engineering, The Chinese University of Hong Kong \and
$^2$ Shenzhen International Graduate School, Tsinghua University
}

\begin{document}
\maketitle
\renewcommand{\thefootnote}{\fnsymbol{footnote}}
% \footnotetext[1]{Accepted in CVPR 2024. This is not the final camera-ready version.}
\footnotetext[2]{Corresponding author.}
\begin{abstract}

Defocus blur is a persistent problem in microscope imaging that poses harm to pathology interpretation and medical intervention in cell microscopy and microscope surgery. To address this problem, a unified framework including the multi-pyramid transformer (MPT) and extended frequency contrastive regularization (EFCR) is proposed to tackle two outstanding challenges in microscopy deblur: longer attention span and data deficiency. The MPT employs an explicit pyramid structure at each network stage that integrates the cross-scale window attention (CSWA), the intra-scale channel attention (ISCA), and the feature-enhancing feed-forward network (FEFN) to capture long-range cross-scale spatial interaction and global channel context. The EFCR addresses the data deficiency problem by exploring latent deblur signals from different frequency bands. It also enables deblur knowledge transfer to learn cross-domain information from extra data, improving deblur performance for labeled and unlabeled data. Extensive experiments and downstream task validation show the framework achieves state-of-the-art performance across multiple datasets. Project page: \href{https://github.com/PieceZhang/MPT-CataBlur}{https://github.com/PieceZhang/MPT-CataBlur}.
 
\end{abstract}

% CSWA: achieve long-range attention modeling with quadratically enlarged equivalent receptive field, image prior \cite{mei_pyramid_2023,michaeli_blind_2014}, efficient \cite{liu_swin_2021}
% ISCA: global attention across feature channels \cite{zamir_restormer_2022}
% FEFN: feature aggregation

  \vspace{-2mm}
\section{Introduction}
\label{sec:intro}

Microscope offers observers enhanced resolution and magnification \cite{ma_comprehensive_2021,cuny_live_2022,masters2008history}, which greatly promotes the advancement of cell microscopy \cite{cuny_live_2022} and surgical microscopy \cite{ma_comprehensive_2021}. 
% It offers the observer enhanced resolution, contrast, and magnification, allowing analysis of the anatomical structure and biological function of small objects that are otherwise invisible to the human eye, 
Cell microscopy employs various optical techniques to reveal the structure and function of cells \cite{cuny_live_2022}.
% Cell imaging employs various optical techniques, such as fluorescence, phase contrast, and confocal, to observe and analyze living or fixed cells and tissues, revealing their structure, function, and dynamics \cite{cuny_live_2022}. 
Surgical microscopy assists surgeons in performing delicate operations \cite{ma_comprehensive_2021} including neurosurgery \cite{yacsargil2013microsurgery}, ophthalmology \cite{bille2019high}, dentistry \cite{garcia2007application}, etc.
% Surgery microscope assists surgeons in performing delicate and precise operations, providing a magnified and illuminated view of the surgical field, enhancing the surgeon’s visibility and accuracy \cite{ma_comprehensive_2021}. It is widely used in clinical practice, such as neurosurgery \cite{yacsargil2013microsurgery}, ophthalmology \cite{bille2019high}, and dentistry \cite{garcia2007application}.
% Microscopy can face technical difficulties that affect imaging quality, such as out-of-focus, noise, spherical aberration, etc. 
In microscopy, out-of-focus, or defocus, is one of the most common visual impairments caused by inferior optical quality, lens aperture, or object magnification \cite{ma_comprehensive_2021,cuny_live_2022}, resulting in blurred or distorted imaging.
It poses harm to the downstream tasks \cite{zhang_benchmarking_2022}, including segmentation \cite{keaton2023celltranspose,kalavakonda2019autonomous}, detection \cite{schmidt_cell_2018}, and classification \cite{chen_automatic_2022}.
% Among them, out-of-focus, or defocus, is one of the most common visual impairments caused by various factors including optical quality,  objective lens aperture, and object magnification \cite{ma_comprehensive_2021,cuny_live_2022}, leading to a blurred or distorted image, which can be disastrous for the downstream tasks \cite{zhang_benchmarking_2022}, such as segmentation \cite{keaton2023celltranspose,kalavakonda2019autonomous}, classification \cite{chen_automatic_2022,ghamsarian_enabling_2020}, and instances counting \cite{riccio2018new,sokolovas2021cell}.
% It can be caused by various factors, such as the optical quality, the ratio of objective lens aperture to the focal length, and the object magnification \cite{ma_comprehensive_2021}. 
While various microscopes with auto-focusing \cite{xu2017wavefront,pinkard2019deep,li2021deep}, assisted-focusing \cite{trukhova2022microlens}, or multi-focus \cite{yoo20183d,lin2019multi} capabilities have been developed to mitigate the defocus effect on-site, 
% as much as possible, 
image degradation remains if the objects are distributed non-uniformly and not co-planar \cite{mazilu_defocus_2023}, or the cavities are too deep to be aligned with the focal plane \cite{ma_comprehensive_2021}. 
% these types usually have higher cost and complex operation with strictly limited application scenarios, 
% This can cause severe visual impairment for cell microscopy when object with thick structures are not co-planar \cite{mazilu_defocus_2023}, as well as for surgical microscopy when the instruments or the cavities in the surgical sites are too deep to be perfectly aligned with the focal plane \cite{ma_comprehensive_2021}. 
Microscopy defocus deblur methods have thus been introduced as an offsite restoration approach.

\begin{figure}[t]
  \centering
  \vspace{-1mm}
   \includegraphics[width=0.85\linewidth]{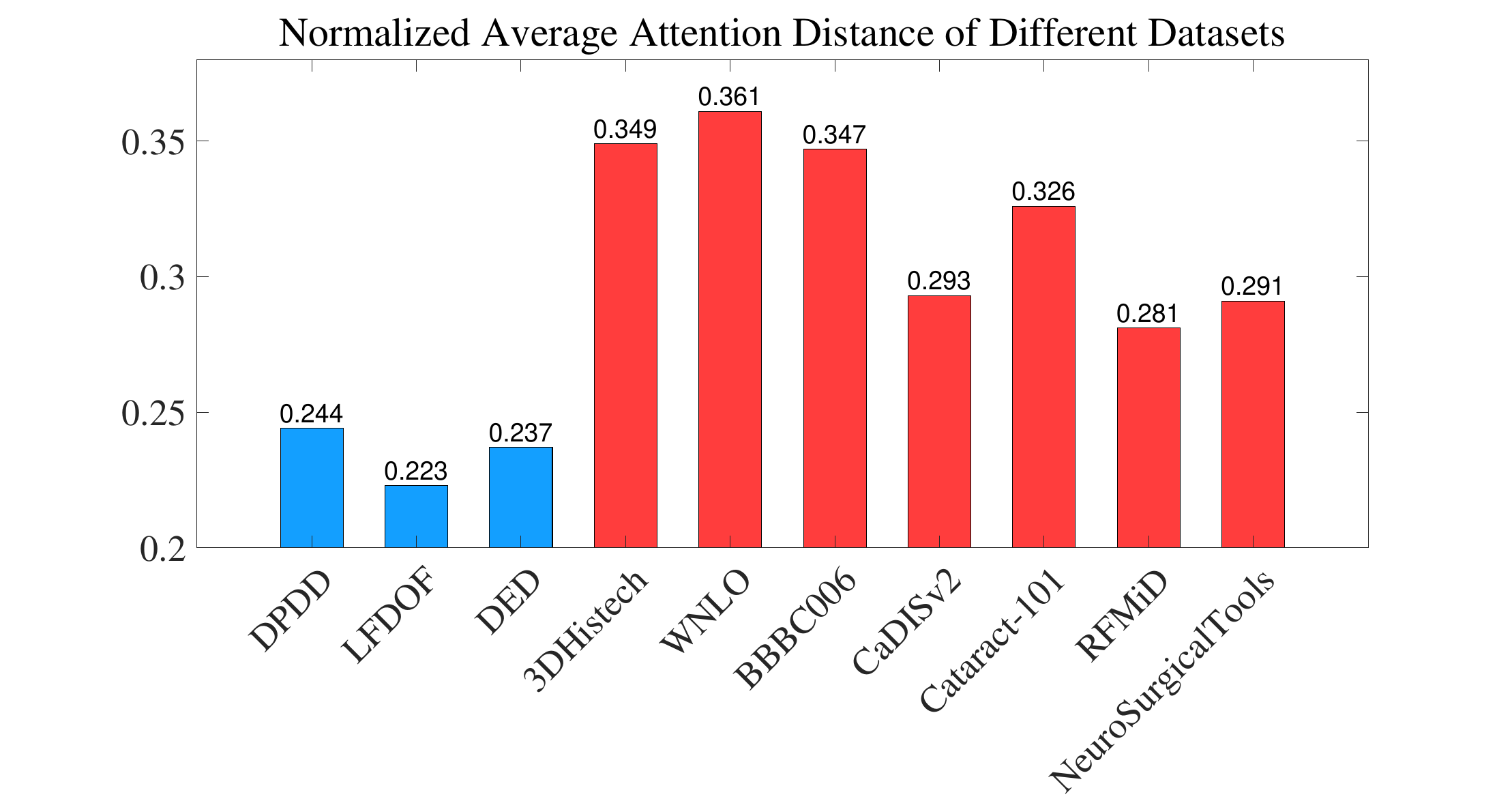}

   \caption{Normalized average attention distance of different datasets. The distance of real-world datasets (shown in blue) is significantly smaller than that of microscopy datasets (shown in red), showing the inter-domain feature difference. \vspace{-3mm}}
   \label{fig:attndist}
  \vspace{-3mm}
\end{figure}
% include more surgical microscope dataset, to prove all surgical microscope image follow similar pattern (like cataract), then justify the reason of using cataract dataset

Recent advances in deep learning have led to the development of various deep defocus deblur methods \cite{ruan_learning_2022,lee_iterative_2021,quan_gaussian_2021,quan_neumann_2023}, including those designed for microscopy \cite{geng_cervical_2022,mazilu_defocus_2023,wang_defocus_2023,zhang_correction_2022,jiang_blind_2020,ghamsarian_deblurring_2020,zhao_new_2020,wang_deblurring_2023,zhang_high-throughput_2023,kornilova_deep_2021,liu_deep_2020}. 
% However, 
Microscopy deblurring poses different challenges from real-world deblur tasks, due to the significant discrepancy between the features in the microscope images and natural scene images \cite{zhang_benchmarking_2022}. 
This difference can be quantified by calculating the normalized average attention distance for different datasets (attention intensity weighted by pixel distance then normalized by image size).
% (for each pixel of an image, multiply attention intensity with other pixels by distance to that pixel, then normalize by image size).
The evaluation involves real-world datasets (DPDD \cite{abuolaim_defocus_2020}, LFDOF \cite{ruan_aifnet_2021}, DED \cite{ma_defocus_2022}), cell microscopy datasets (3DHistech \cite{geng_cervical_2022}, WNLO \cite{geng_cervical_2022}, BBBC006 \cite{ljosa_annotated_2012}), and surgical microscopy datasets for cataract surgery (CaDISv2 \cite{grammatikopoulou_cadis_2022}, Cataract-101 \cite{schoeffmann_cataract-101_2018}), retinal microsurgery (RFMiD \cite{pachade_retinal_2021}), neurosurgery (NeuroSurgicalTools \cite{bouget_detecting_2015}). 
As shown in \cref{fig:attndist}, all cell microscopy datasets have normalized average attention distance around 0.35, and surgical microscope datasets around 0.3, indicating a much longer attention span than the real-world datasets at under 0.25. 
This result reveals the substantial discrepancy between these two domains, suggesting that \textbf{modeling attention in wider areas with larger receptive fields would benefit microscopy tasks}.
% This implies that long-range modeling strategies could benefit microscope image tasks, and attention could be based on similar feature patches that recur at distant positions \cite{mei_pyramid_2023}. 
% Several existing microscope defocus deblur methods implicitly model long-range correspondences by using coarse-to-fine structures \cite{wang_defocus_2023,liu2022coarse,ghamsarian_deblurring_2020} or multi-scale features \cite{geng_cervical_2022}. However, these implicit approaches are limited by the filter size and network structure, resulting in marginal improvement in long-range modelling. 
% The idea of long-range modelling has also been applied to many other low-level tasks \cite{cho_rethinking_2021,ruan_learning_2022,zhang_dynamic_2022,mei_pyramid_2023,gendy_balanced_2022,michaeli_blind_2014}. 
% Among them, methods based on self-similarity prior have been extensively explored and proven to be effective in many related works \cite{mei_pyramid_2023,michaeli_blind_2014,zha2020image,yu2017blind,liu2019nonblind}, including microscope \cite{komiyama2022three,napoletano2018anomaly} and medical \cite{liu2019medical,xu2019self} applications. 
% Self-similarity refers to the recurrence of small similar patches at different positions and scales, which provide clean feature correspondences in multi-scale image pyramid \cite{mei_pyramid_2023}, as corruptions decrease significantly at coarser image scales \cite{zontak2013separating,michaeli_blind_2014}. 
Motivated by this analysis, we introduce a multi-pyramid transformer (MPT) with cross-scale window attention (CSWA), intra-scale channel attention (ISCA), and feature-enhancing feed-forward network (FEFN), to construct multiple pyramids explicitly on each stage of the network, fully exploiting latent cross-scale features in every projection space.
CSWA captures the interaction between local $query$ and cross-scale $key$-$value$ pairs for long-range attention modeling with a quadratically enlarged receptive field while keeping computational efficiency.
% relying on the cross-scale attention distribution similarity \cite{li_efficient_2023}, while keeping efficiency by employing window attention \cite{liu_swin_2021} instead of global spatial attention \cite{dosovitskiy_image_2021}.
% CSWA can also benefit from the significant corruption drop occurred at coarser scales \cite{zontak2013separating,michaeli_blind_2014}.
ISCA builds channel-wise attention on a local scale to provide global channel context, which is then integrated with the spatially correlated feature from CSWA by the proposed FEFN through an asymmetric activation mechanism.
% MPT constructs multiple pyramids explicitly on each stage of the network, fully exploiting latent cross-scale features in every projection space. 
% which enables long-range modeling and balance between performance and complexity.

Another problem in microscopy deblur is the insufficient data for training a robust model.
Different from natural scene defocus deblur methods that use datasets captured with varying aperture sizes \cite{abuolaim_defocus_2020} or light field camera \cite{ruan_learning_2022,ma_defocus_2022}, the high-quality training data for microscopy deblur can be much harder to obtain\cite{geng_cervical_2022,ma_comprehensive_2021}.
% Natural scene defocus deblur methods often use datasets captured with different aperture sizes to obtain all-in-focus and defocused image pairs \cite{abuolaim_defocus_2020}, or synthesize defocused images from light field camera datasets \cite{ruan_learning_2022,ruan_aifnet_2021,ma_defocus_2022}. 
% For cell microscopy, the feature in training set is usually insufficient for training a generalizable model, since there is a huge feature discrepancy between training domain and unseen domain caused by various staining and imaging methods \cite{geng_cervical_2022,zhang_benchmarking_2022}.
For cell microscopy, insufficient training feature leads to a generalizability problem caused by different staining and imaging methods \cite{geng_cervical_2022,zhang_benchmarking_2022}.
% all-in-focus and defocused training pairs can be acquired by manually adjusting the focal plane \cite{geng_cervical_2022,ljosa_annotated_2012}, 
% The situation is worse for surgical microscopy as no labeled dataset is available because data acquisition will undoubtedly hinder the normal surgical procedure \cite{ghamsarian_deblurring_2020,ma_comprehensive_2021}.
The situation is worse for surgical microscopy because the imaging principle of microscope makes it impossible to simultaneously acquire blur-sharp pairs for model training \cite{ghamsarian_deblurring_2020,ma_comprehensive_2021}.
% {\color{blue}This does not sound convincing. Why would data acquisition hinder the surgical procedure?}
% Previous works address this by training with dataset synthesized using uniform Gaussian kernel \cite{ghamsarian_deblurring_2020}, inevitably leading to feature discrepancy with real blur type. 
To alleviate the data deficiency problem, some training diagrams learn rich information by extending extra training data and then fine-tuning it with testing data \cite{ruan_learning_2022}. 
This paradigm, however, may not be applicable to microscopy deblurring, as there is an \textbf{inter-domain gap} between natural scenes and microscopy images, and also \textbf{intra-domain gaps} among different microscopy datasets that are highly task-specific.
% microscopy dataset is highly task-specific leading to huge intra-domain gaps between its counterparts hindering effective knowledge transfer, not to mention the existing data deficiency problem. 
The extended frequency contrastive regularization (EFCR) is proposed to address the data deficiency problem by encouraging the model to learn representations from decoupled frequency bands in the wavelet domain \cite{zhang2019wavelet,he2023camouflaged}, and further exploiting latent information leveraging the fact that model trained with synthetic reblurring images can deblur its naturally blurred counterpart \cite{ghamsarian_deblurring_2020}. 
It also enables cross-domain deblur knowledge transfer, facilitating multiple scenarios including extra data training and unlabeled data deblur.

% Apart from adopting frequency directly for comparison \cite{bai_contrastive_2022}, namely absolute contrastive, the novel relative contrastive learning is proposed by taking the frequency residual for comparison to enforce the model learning latent deblur guidance beyond the pixel-wise constraint.
% Since frequency residual is data-agnostic, it also enables cross-domain knowledge transferring.

% A novel realistic blur synthesizing method is proposed in this paper with a new defocus dataset collected in microscope surgery to prove training with realistic blur does benefit deblur outcome.
This paper presents a unified deblur framework with MPT and EFCR to address the aforementioned two challenges in microscopy deblur. 
The surgical microscopy deblur is illustrated on cataract surgery, which is the most common surgery worldwide \cite{trikha2013journey,fang2022global,han2023real}.
% {\color{blue}I think we should keep this sentence.} 
Extensive experiments are conducted on various open-source cell and surgical microscopy datasets, along with downstream tasks validation on cell detection and surgery scene semantic segmentation.
For surgical microscopy deblur, we present a 
% \sout{novel} {\color{blue}I dont think you need to call it novel here as it is not being considered as one of your key contributions.} 
realistic blur synthesizing method, and collect a new dataset of defocus cataract microscopic surgery, which is the first dataset for surgical microscopy deblur. 
% Extensive experiments are conducted on three cell microscopy dataset \cite{geng_cervical_2022,ljosa_annotated_2012} and three surgical microscopy dataset \cite{schoeffmann_cataract-101_2018,grammatikopoulou_cadis_2022}, along with a natural scene dataset \cite{abuolaim_defocus_2020} .
The method achieves state-of-the-art performance on not only microscope datasets but also real-world datasets, showing the universality of the proposed framework. 
The deblur results on unlabeled datasets also prove the effectiveness of the proposed EFCR on knowledge transferring.
The main contributions are as follows,
% {\color{blue}I think you can add some brief sentences about the impact or significance for your contribution 1 and 2 below to show how these proposed methods differentiate themselves from existing deblurring methods or tackle some fundamental issues that existing methods cannot resolve which therefore lead to the better deblurring outcome. The main idea is that your contributions 1 and 2 in their current forms only look like an iteration of your proposed method without informing the readers how they are better than the state-of-the-art. Many readers would not read your whole introduction carefully to see the detailed limitations of existing methods, so you somehow need to very concisely highlight it here.}: 
% 1) A novel multi-pyramid transformer (MPT) is proposed to address defocus blur in microscopy. {\color{blue}This sentence in itself cannot be a contribution since MPT is not new and has been adopted for this task too.}
1) The multi-pyramid transformer (\textbf{MPT}) is for the first time proposed for microscopy defocus deblur.
% {\color{blue}Yuelin, is MPT proposed for the first time for debluring? Or MPT+CSWA is proposed for the first time for debluring. Reply: Actually both of them.}
It models the long-range spatial attention between local-scale and down-scale maps in each explicit pyramid using the proposed cross-scale window attention (\textbf{CSWA}) with a quadratically enlarged receptive field to adapt to the longer attention span of microscopy datasets. 
2) The intra-scale channel attention (\textbf{ISCA}) is presented to incorporate global channel context in the CSWA spatial information via the proposed feature-enhancing feed-forward network (\textbf{FEFN}), providing additional intra-scale channel features to the pyramid.
3) A training strategy with extended frequency contrastive regularization (\textbf{EFCR}) is presented to alleviate data deficiency by exploiting latent deblur signal beyond the pixel constraint through synthetic reblurring, which is the first implementation of contrastive learning in microscopy deblur. It also enables cross-domain deblur knowledge transfer, facilitating extra data training and enhancing unlabeled image deblur.

\section{Related Work}
\label{sec:relatedwork}
% {\color{blue}A general comment on this section is that you provide the limitations of existing methods so clearly that when readers read your contributions statements above or the detailed methods section, they would immediately know you have address these current limitations. So I would say if you think you have explained well the limitations of existing methods (but make sure you really do explain well), then you do not have to state your proposed work and how it solves the limitations in this section. Then that would help to save some space. }

%-------------------------------------------------------------------------
\begin{figure*}[!ht]
  \centering
   %\includegraphics[width=0.8\linewidth]{egfigure.eps}
   \includegraphics[width=1\linewidth]{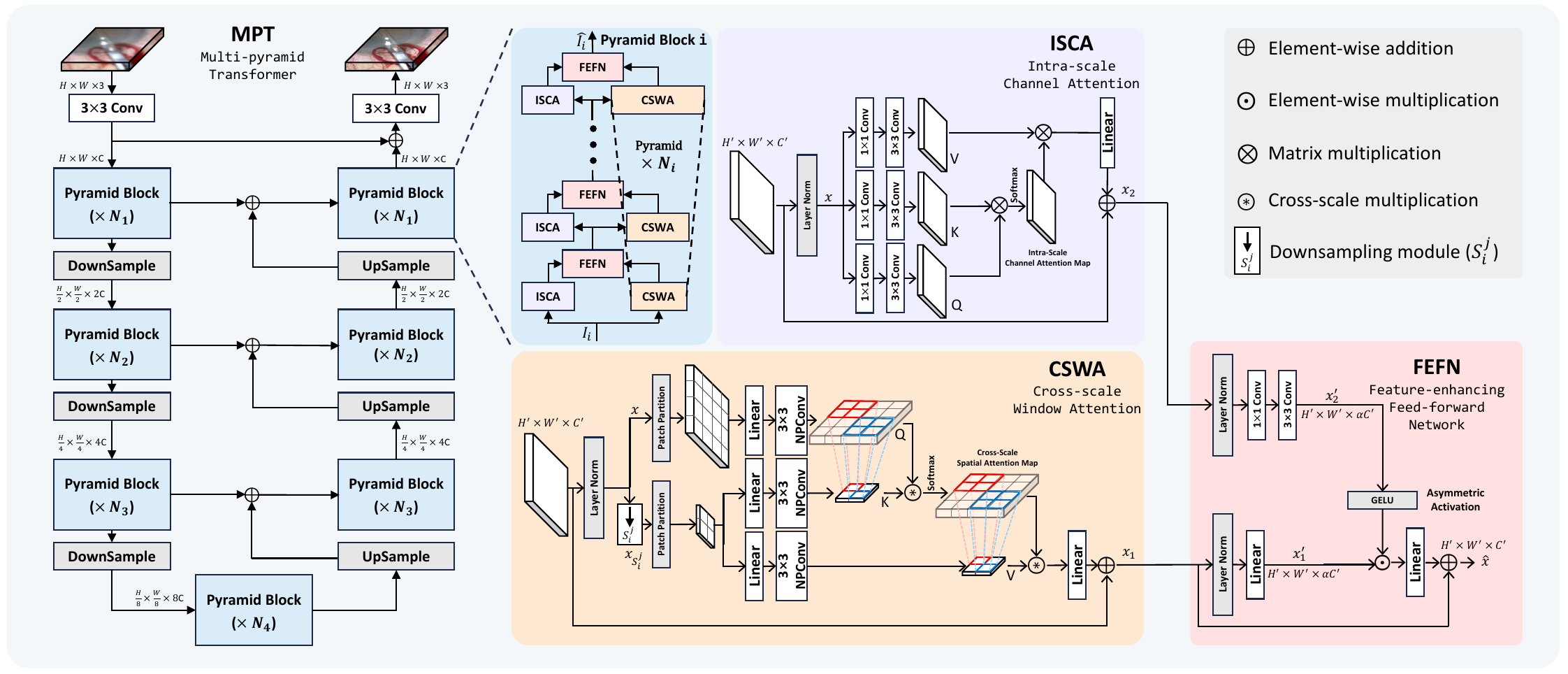}
   \caption{Overview of MPT. MPT constructs an explicit pyramid block at each stage. Inside the pyramid block, CSWAs constitute a coarse-to-fine pyramid, exploring cross-scale spatial interaction for each scale. The ISCA is built beside each CSWA to provide global channel context. Information from CSWA and ISCA is aggregated by FEFN using the asymmetric activation mechanism.\vspace{-4mm}}
   \label{fig:MPT}
\end{figure*}

\paragraph{Single image defocus deblurring}
% Defocus blur with non-uniform spatially variant characteristics is difficult to restore.
% Learning-based methods achieve better results than traditional hand-crafted feature-based methods \cite{cho2017convergence,karaali2017edge,shi2015just} thanks to its ability of high dimensional representation modeling. 
% Many efforts have been done on deblurring methods based on defocus map estimation (DME) \cite{lee_deep_2019,ma_defocus_2022,xin_defocus_2021,ruan_aifnet_2021,lee_iterative_2021}, but failed to popularize it to microscopy deblur due to the obstacles on extra training data \cite{ma_defocus_2022,lee_deep_2019} and robustness \cite{ruan_learning_2022}.
% DME methods restore the defocused image by leveraging information from estimated defocus map, which containing the size of circle of confusion (COC) per pixel \cite{potmesil1981lens}. 
% However, it has been shown that this intermediate procedure does not always benefits the restoration performance since accurate defocus map cannot be guaranteed \cite{ruan_learning_2022}, although extra training signal from defocus map label is introduced \cite{ma_defocus_2022,lee_deep_2019}. These shortcoming hindering DME methods from being applied in microscope deblurring. 
For learning-based deblur model, end-to-end method is widely applied \cite{lee_iterative_2021,son_single_2021,ruan_learning_2022,quan_neumann_2023,quan_gaussian_2021} for its better performance and robustness \cite{quan_gaussian_2021} than methods based on defocus map estimation \cite{lee_deep_2019,ma_defocus_2022,ruan_aifnet_2021,lee_iterative_2021}. 
% and hand-crafted feature \cite{cho2017convergence,karaali2017edge,shi2015just}.
% including many microscope deblurring works \cite{geng_cervical_2022,mazilu_defocus_2023,wang_defocus_2023,zhang_correction_2022,jiang_blind_2020,ghamsarian_deblurring_2020,zhao_new_2020,wang_deblurring_2023,zhang_high-throughput_2023,kornilova_deep_2021,liu_deep_2020}. 
% Since these methods keep training goal more consistent than DME methods, they usually achieve higher accuracy and robustness \cite{quan_gaussian_2021}. 
Among them, many deblurring works have been done on cell microscopy to alleviate defocus brought by mechanical axial shift \cite{wang_defocus_2023} and non-coplanar cells \cite{mazilu_defocus_2023}, and enhance the imaging quality of human cell \cite{mazilu_defocus_2023,wang_deblurring_2023}, pathology \cite{geng_cervical_2022,jiang_blind_2020}, parasite \cite{zhang_correction_2022}, etc. 
% {\color{red}A blur synthesizing method was presented to simulate realistic blur in surgical microscopy, addressing the problem in the existing blur synthesizing method \cite{ghamsarian_deblurring_2020}. 
% A generalizable unified microscope deblur model MPT {\color{blue}Is this not a multi-scale method?} is proposed to handle multi-type blur in various microscope tasks.}
Compared with cell microscopy, surgical microscopy deblur has not been well-explored due to the difficult acquisition of blur datasets with ground truth
% the nature of microscopic surgery that blur dataset with ground truth is hard to obtain 
~\cite{ghamsarian_deblurring_2020}, although defocus blur is commonly encountered in microscopic surgery \cite{ghamsarian_deblurring_2020,zhao_new_2020}.

\vspace{-3mm}
\paragraph{Multi-scale methods}  % method to enlarge receptive field
% Multi-scale structure with image pyramid is widely adopted in neural networks to enable multi-scale information aggregation and enlarge receptive field.
The image pyramid in existing methods is usually built in two ways, i.e., explicitly stacking multi-scale feature maps in a single pyramid \cite{mei_pyramid_2023,gendy_balanced_2022,xu_edpn_2021}, or implicitly applying multi-stage structure \cite{cho_rethinking_2021,kim_mssnet_2023,ruan_learning_2022,cui_dual-domain_2023,quan_gaussian_2021,wang_pyramid_2021}. 
% These design paradigms have been widely applied in many microscopy deblur methods \cite{geng_cervical_2022,wang_defocus_2023,jiang_blind_2020,ghamsarian_deblurring_2020}. 
% Explicit pyramid methods build cross-scale attention in the pyramid composed of stacked downscaled maps \cite{mei_pyramid_2023,gendy_balanced_2022,xu_edpn_2021}, causing single-level feature deficiency since the explicit pyramid is built by downscaling feature in a single latent space.
Explicit pyramid methods face single-level feature deficiency since the explicit pyramid is built on downscaling features in a single latent space \cite{mei_pyramid_2023,gendy_balanced_2022,xu_edpn_2021}.
% cannot make full use of the projection space across the whole model.
% Some of them globally model whole-pyramid attention but lead to computationally intensive quadratic complexity \cite{gendy_balanced_2022,mei_pyramid_2023}.
% Implicit pyramid methods, or multi-stage methods, perform channel-level or spatial-level feature aggregation at each gradually downscaled stage \cite{zamir_restormer_2022,cho_rethinking_2021,cui_dual-domain_2023,ruan_learning_2022,wang_pyramid_2021,geng_cervical_2022,jiang_blind_2020,wang_defocus_2023,ghamsarian_deblurring_2020} or apply multi-scale input or output  \cite{cho_rethinking_2021,kim_mssnet_2023,cui_dual-domain_2023,quan_gaussian_2021,wang_defocus_2023,ghamsarian_deblurring_2020}. 
% Most of the existing microscope defocus deblur methods adopt multi-stage structures with implicit pyramids to capture multi-scale correlation thus enlarging receptive field implicitly \cite{wang_defocus_2023,liu2022coarse,ghamsarian_deblurring_2020,geng_cervical_2022}.  
Most of the existing microscope deblur methods \cite{wang_defocus_2023,liu2022coarse,ghamsarian_deblurring_2020,geng_cervical_2022} adopt implicit multi-stage design to perform aggregation on separated latent space but suffer from inter-level feature discrepancy. 
% The receptive field is also confined by the network structure, resulting in marginal improvement.
The proposed MPT in this work addresses these drawbacks by building \textbf{multiple explicit pyramids} with CSWA, ISCA, and FEFN on\textbf{ each feature level}, achieving cross-scale feature aggregation with an enlarged receptive field.

\vspace{-4mm}
\paragraph{Contrastive learning}
% Except for being widely adopted in unsupervised representation learning by instance discrimination \cite{he_momentum_2020,chen_simple_2020,pielawski_comir_2020}, 
Contrastive learning has been widely applied in low-level tasks \cite{jan_contrastive_2022,zhao_real-aware_2023,wu_practical_2023,feng_hierarchical_2023,bai_contrastive_2022,zhao_fcl-gan_2022,he2024strategic}. 
They construct contrastive pairs on the feature space that take clean and corrupted images as positive and negative pairs, respectively \cite{wu2021contrastive,liang2022semantically}.  
% \sout{Beyond the pixel, some works have been done leveraging contrastive frequency information \cite{bai_contrastive_2022,wu_practical_2023,zhao_fcl-gan_2022} and many of them extract frequency representations by wavelet transform \cite{zhang2019wavelet,bai_contrastive_2022,wu_practical_2023}. }
Some works have leveraged contrastive frequency information \cite{zhao_fcl-gan_2022} and extracted frequency representations by wavelet transformation~\cite{zhang2019wavelet,bai_contrastive_2022,wu_practical_2023}. 
Following the fact that sharp and blurry images have similar low-frequency components but differ significantly in the high-frequency part \cite{wu_practical_2023}, the idea of comparing different frequency bands separately is adopted in \cite{zhang2023multi} by applying $\mathcal{L}_1$ loss directly to the frequency bands in contrastive regularization. 
In this paper, the proposed EFCR adopts basic CR and extended CR to encourage learning latent deblur signals and transferring cross-domain deblur information, thus addressing the data deficiency problem.
% To the best of our knowledge, this is the first implementation of contrastive learning in microscopy deblur. {\color{blue}I think this last sentence should be moved to your contribution 3.}

\vspace{-2mm}
\section{Method}
\label{sec:method}

The proposed framework consists of MPT and EFCR to address the two outstanding problems in microscopy defocus deblur, namely longer attention span and data deficiency. 
% This section gives the details about MPT with CSWA, ISCA, and FEFN mechanism, and EFCR.

%-------------------------------------------------------------------------
% \begin{figure*}[t]
%   \centering
%   % \fbox{\rule{0pt}{2in} \rule{0.9\linewidth}{0pt}}
%    %\includegraphics[width=0.8\linewidth]{egfigure.eps}
%    \includegraphics[width=1\linewidth]{fig/mpt.pdf}
%    \caption{Overview of MPT. MPT constructs an explicit pyramid block at each stage. Inside the pyramid block, CSWAs constitute a coarse-to-fine pyramid, exploring cross-scale spatial interaction for each scale. The ISCA is built along each CSWA to provide global channel context. Information from CSWA and ISCA is aggregated by FEFN using the asymmetric activation mechanism.}
%    \label{fig:MPT}
% \end{figure*}

\subsection{Multi-pyramid Transformer (MPT)}
\label{sec:methodMPT}
MPT builds multiple explicit pyramids on each feature level, thus avoiding single-level feature deficiency and inter-level feature discrepancy \cite{mei_pyramid_2023,xu_edpn_2021,wang_defocus_2023,ghamsarian_deblurring_2020}.
As shown in \cref{fig:MPT}, the proposed MPT follows a U-shaped structure \cite{ronneberger_u-net_2015,wang_uformer_2022,he2023retidiff}.
The blur image $I_{in}$ with the size of $H \times W$ first gets input feature projection $I_1\in\mathbb{R}^{H\times W\times C}$ through a convolution.  
Then, the feature goes through the network with seven pyramid blocks followed by re-sampling using pixel shuffles \cite{shi_real-time_2016}, and finally projects back to the image with a convolution.
% After passing the latent pyramid stage, the feature $I_4\in\mathbb{R}^{\frac{H}{8}\times \frac{W}{8} \times 8C}$ proceeds to the decoder with three pyramid blocks cascaded after pixel-unshuffle upsampling \cite{shi_real-time_2016}, and finally projects back to image space with a convolution.
% $I_{out}\in\mathbb{R}^{H\times W\times 3}$
The shortcut connection is built between each encoder and decoder stage by element-wise addition.
% Image pyramids in MPT are built explicitly in pyramid blocks on every feature levels, thus avoiding single-level feature deficiency and inter-level feature discrepancy.

\vspace{-3mm}
\paragraph{Pyramid block}
The pyramid block $i$ receives $I_i\in\mathbb{R}^{\frac{H}{2^i}\times \frac{W}{2^i} \times 2^iC}$ ($i \in \{ 1,2,3,4\}$), and outputs the map $\hat{I_i}$ in the same size. 
There are $N_i$ sub-blocks in pyramid block $i$. Each of them handles a local scale $S_i^j$ ($j\in \{ 1, 2, ..., N_i \}$, $ S_i^j \in \{\frac{1}{8}, \frac{1}{4}, \frac{1}{2}, 1 \}$).
Multiple stacked sub-blocks build the pyramid in a coarse-to-fine manner, i.e.,  $S_i^k \leq S_i^{k+1}$, $ 1 \leq k < N_i $.
This design achieves progressive multi-scale feature aggregation and ensures the full exploration of each scale.
In practice, $N_i$ is set to be an even number. The sub-block in the even level adopts the common cyclic window shifting strategy  \cite{liu_swin_2021} to gain cross-window interaction.

% CSWA: achieve long-range attention modeling with quadratically enlarged equivalent receptive field, image prior \cite{mei_pyramid_2023,michaeli_blind_2014}, efficient \cite{liu_swin_2021}
\vspace{-3mm}
\paragraph{Cross-scale window attention (CSWA)}
CSWA captures the long-range interaction by modeling the attention between windows from the local-scale map and the down-scale map. 
The layer normalized \cite{ba2016layer} feature $x\in\mathbb{R}^{H^{\prime}\times W^{\prime}\times C^{\prime}}$ first passes through a downsampling module to get $x_{S_i^j}$ with the size of ${H^{\prime}S_i^j\times {W^{\prime}}{S_i^j} \times C^{\prime}}$. 
% where $H^{\prime}$, $W^{\prime}$, and $C^{\prime}$ is the map size of pyramid block $i$. 
A strided average pooling followed by a linear projection with a shortcut is adopted in the downsampling module.
% as a trainable projection to alleviate information loss brought by downscaling.
% Then, $x$ and $x_{S_i^j}$ are segmented into $\frac{H^{\prime}W^{\prime}}{M^2}$ and $\frac{H^{\prime}W^{\prime}({S_i^j})^2}{M^2}$ non-overlapping patches in the size of $M\times M$.
% The windowed self-attention \cite{liu_swin_2021} is calculated by non-overlapped segmented $x$ and $x_{S_i^j}$ to get \textit{query} ($Q\in \frac{H^{\prime}W^{\prime}}{M^2} \times M^2 \times C^{\prime}$) and \textit{key-value} ($K,V\in \frac{H^{\prime}W^{\prime}({S_i^j})^2}{M^2} \times M^2 \times C^{\prime}$) pair, where $M$ is the patch width.
% by applying a linear projection followed by a $3\times 3$ neighboring padding convolution (NPConv). 
The $3\times 3$ neighboring padding convolution (NPConv) is proposed to generate $Q, K, V$ projection ($Q\in \mathbb{R}^{\frac{H^{\prime}W^{\prime}}{M^2} \times M^2 \times C^{\prime}}$, $K,V\in \mathbb{R}^{\frac{H^{\prime}W^{\prime}({S_i^j})^2}{M^2} \times M^2 \times C^{\prime}}$) with inductive bias\cite{wu_cvt_2021,guo_cmt_2022}, where $M$ is the patch width. It pads a patch with its neighborhood pixels, providing the isolated edge pixels with neighboring information \cite{vaswani_scaling_2021} (all $3\times 3$ convolutions adopt bias-free grouped convolution by default with group size equal to the feature dimension).
%%%%%%%%%%%%
To get the cross-scale spatial attention map, the cross-scale multiplication ($\circledast$) is introduced as a one-to-many strategy, where each patch in $K$ is multiplied with $\frac{1}{S_i^j}\times \frac{1}{S_i^j}$ patches in the corresponding location of $Q$, as illustrated in \cref{fig:MPT}.
This operation provides an $M\times M$ local patch in $Q$ with interaction with a $M\times M$ patch in $K$ whose information comes from a $\frac{M}{S_i^j}\times \frac{M}{S_i^j}$ region in the original map.
Although the downscaled map loses information, it can still be highly instructive, since attention distributions of different scales are highly consistent \cite{li_efficient_2023}. It also leverages the pool of sharper patches generated by downscaling which serve as \textit{priors} for deblurring \cite{zontak2013separating,michaeli_blind_2014}.
It makes the receptive field quadratically enlarged by $(S_i^j)^2$ times while keeping the computational complexity $O(M^2H^{\prime}W^{\prime}C^{\prime})$ unchanged as the vanilla local window attention \cite{liu_swin_2021}.
% $O(\frac{H^{\prime}W^{\prime}}{M^2}M^4C^{\prime})=O(M^2H^{\prime}W^{\prime}C^{\prime})$
% It also keeps the positional correspondence since the cross-scale attention distribution shows high similarity \cite{li_efficient_2023}.
A similar strategy is adopted in multiplying the attention map and $V$.
%%%%%%%%%%%%
The self-attention in CSWA for a local window in the size of $M^2\times C$ can be defined as:
\vspace{-2mm}
\begin{equation}
  Attention_1(q,k,v)=Softmax(qk^T/\sqrt{d}+B)v,  % \circledast
  \label{eq:cswa}
  \vspace{-2mm}
\end{equation}
where $q,k,v\in \mathbb{R}^{M^2\times d}$, and $B$ is the relative positional encoding \cite{liu_swin_2021}. 
In practice, we implement the multi-head self-attention \cite{vaswani_attention_2017} by concatenating the result of $h$ parallelly calculated attention.
The output $x_1$ is then obtained by a linear projection with a shortcut.

% ISCA: global attention across feature channels \cite{zamir_restormer_2022}
\vspace{-3mm}
\paragraph{Intra-scale channel attention (ISCA)} 
ISCA handles a single scale feature $x\in\mathbb{R}^{H^{\prime}\times W^{\prime}\times C^{\prime}}$ and generates cross-channel interaction with encoded global context \cite{zamir_restormer_2022}.
By applying $1\times 1$ convolutions followed by $3\times 3$ convolutions, projections $Q,K,V\in \mathbb{R}^{{H^{\prime}W^{\prime}\times C^{\prime}}}$ are generated from layer normalized $x$, introducing convolutional inductive bias and extracting cross-channel information in both point-wise and spatial-wise manners.
The intra-scale channel attention map is then calculated by multiplying $Q$ with $K$. The self-attention in ISCA can be formulated as:
\vspace{-2mm}
\begin{equation}
  Attention_2(Q,K,V)=Softmax(QK^T)V
  \label{eq:cswa}
  \vspace{-2mm}
\end{equation}
Similar to CSWA, ISCA implements the multi-head self-attention \cite{vaswani_attention_2017} to get $x_2$.
% The self-attention result finally goes through a linear projection with a shortcut and gets $x_2$, similar to CSWA.
% The channel context from ISCA will function as a feature complement in the following feed-forward network. 

% FEFN: feature aggregation
\vspace{-3mm}
\paragraph{Feature-enhancing feed-forward network (FEFN)}
The FEFN aggregates the spatial-wise feature $x_1$ with the channel-wise context $x_2$.
The input features are first projected to $x_1^{\prime}$ and $x_2^{\prime}$ in the size of  $H^{\prime}\times W^{\prime}\times \alpha C^{\prime}$, where $\alpha$ is the expansion ratio.
Instead of combining  $x_1^{\prime}$ and $x_2^{\prime}$ by simply adding them together like \cite{li_efficient_2023}, FEFN adopts an asymmetrical activation mechanism with GELU \cite{hendrycks2016gaussian}, where $x_1$ is element-wisely multiplied by GELU activated $x_2$. 
The FEFN can be formulated as
\vspace{-2mm}
\begin{equation}
  \hat{x}=W_p(GELU(x_2^{\prime}) \odot x_1^{\prime}) + x_1,
  \label{eq:cswa}
  \vspace{-2mm}
\end{equation}
where $\hat{x}\in \mathbb{R}^{H^{\prime}\times W^{\prime}\times C^{\prime}}$, and $W_p$ refers to linear projection.
Compared with the regular FN \cite{dosovitskiy_image_2021}, this asymmetrical operation allows the spatial information from $x_1$ to be guided by the non-linearly activated signal from the channel context in $x_2$, offering $x_1$ an extra global view in terms of feature channels.

\subsection{Extended Frequency Contrastive Regularization (EFCR)}
\label{sec:methodEFCR}

The proposed EFCR contains basic CR and extended CR to explore latent deblur guidance beyond pixel constraints.
% \begin{figure}[t]
%   \centering
%   \fbox{\rule{0pt}{2in} \rule{0.9\linewidth}{0pt}}
%    %\includegraphics[width=0.8\linewidth]{egfigure.eps}

%    \caption{Extended CR in EFCR.}
%    \label{fig:EFCR}
% \end{figure}

\vspace{-3mm}
\paragraph{Constructions of contrastive pairs}
Given a training pair $i$ with ground truth $I_i^{gt}$, blur input $I_i^{in}$, and deblurred output $I_i^{out}$, the Haar wavelet transformation \cite{zhang2019wavelet} decouples the samples into low-low (LL), low-high (LH), high-low (HL), and high-high (HH) bands. 
For simplicity, here we define $f^{h}(\cdot)$ as the operator decoupling and concatenating high-frequency bands (LH, HL, HH), and $f^{l}(\cdot)$ as the operator for low-frequency band (LL).
For basic CR, the frequency bands are directly taken as contrastive pairs.
The positive and negative basic CR  $\mathcal{L}_i^{+}$ and  $\mathcal{L}_i^{-}$ are given by:
% \vspace{-1mm}
\begin{align}
      & \resizebox{0.43\textwidth}{!}{$\mathcal{L}_i^{+}= {\parallel}f^h(I_i^{out})-f^h(I_i^{gt}){\parallel}_1 + {\parallel}f^l(I_i^{out})-f^l(I_i^{gt}){\parallel}_1,$} \label{eq:posCR} \\
      & \resizebox{0.25\textwidth}{!}{$\mathcal{L}_i^{-}= {\parallel}f^h(I_i^{out})-f^h(I_i^{in}){\parallel}_1.$}
  \label{eq:negCR}
  \vspace{-2mm}
\end{align}
% where $I_i^{gt}$,  $I_i^{in}$, and  $I_i^{out}$ are taken as positive, negative, and anchor samples, respectively.
Both bands are included in $\mathcal{L}_i^{+}$, since both high and low frequencies of $I_i^{out}$ are expected to be pulled closer to $I_i^{gt}$.
Only high frequency is taken for $\mathcal{L}_i^{-}$ to push the $f^h(I_i^{out})$ away from $f^h(I_i^{in})$ as blur degradation mainly happens in the high-frequency parts \cite{cui_dual-domain_2023,liu_motion_2020}.

The extended CR enforces the model to learn latent information from degraded high-frequency components beyond the pixel-wise constraint. 
Based on the idea that a model trained with synthetic blurred images can deblur natural blurry images in the dataset \cite{ghamsarian_deblurring_2020}, the blurred image $B_i^{in}$ is generated by applying random Gaussian blur (kernel size in $\{3,5,7\}$) on $I_i^{in}$, followed by calculating its deblurred result $B_i^{out}$.
The extended CR $\mathcal{L}_i^{ext}$ based on extended training pair ($B_i^{in}, B_i^{out}$) is then formulated as:
% \vspace{-2mm}
\begin{equation}
  \mathcal{L}_i^{ext}= \frac{{\parallel}f^h(B_i^{out})-f^h(B_i^{in}){\parallel}_1}{{\parallel}f^h(I_i^{in})-f^h(B_i^{in}){\parallel}_1}.
  \label{eq:extCR}
  \vspace{-2mm}
\end{equation}
% The $\mathcal{L}_1$ loss between $S_i^{out}$ and $S_i^{in}$ is then derived and normalized with the $\mathcal{L}_1$ distance between high frequency part of $I_i^{in}$ and its blurred counterpart $S_i^{in}$.
The $\mathcal{L}_i^{ext}$ is derived as a relative loss term by normalizing with $\mathcal{L}_1$ distance between the high-frequency components of $I_i^{in}$ and its blurred counterpart $B_i^{in}$ to alleviate the disturbance caused by blur variance from 3D objects with different depths \cite{dong_no-reference_2019}.
% the disturbance from the gradient distribution of reblurred image \cite{dong_no-reference_2019}.

% By normalizing the gradient distribution difference, the extended CR reduces the impact from the original image’s blur degree and makes the loss term relative rather than absolute, thus enhancing the robustness of extended CR. 
% This is based on the observation that a stronger blur in the original image tends to result in a smaller distance between it and its reblurred counterpart, and vice versa \cite{dong_no-reference_2019}.

The overall optimization objective $\mathcal{L}$ with the proposed EFCR $\mathcal{L}_{CR}$ is given by:
\begin{equation}
  \mathcal{L}= \mathcal{L}_1 + \beta \mathcal{L}_{CR} = \mathcal{L}_1 +\beta \frac{1}{n}\sum_{i=1}^n\frac{\mathcal{L}_i^{+}}{\mathcal{L}_i^{-}+\mathcal{L}_i^{ext}},
  \label{eq:overallCR}
  \vspace{-2mm}
\end{equation}
where $\mathcal{L}_1$ is the supervised pixel loss, $n$ is the number of samples, and $\beta$ is the scaling factor.

\vspace{-3mm}
\paragraph{Knowledge transfer from extra data}
Defocus blur mainly causes high-frequency degradation \cite{cui_dual-domain_2023,liu_motion_2020}, implying that the high-frequency part can provide informative cross-domain deblur guidance.
EFCR with extra data (denoted by EFCR$_{ex}$) constructs contrastive pairs on high-frequency components.
Given an extra training pair $\{ I_i^{gt^{\prime}}, I_i^{in^{\prime}}, I_i^{out^{\prime}} \}$ from external dataset and corresponding extended samples $\{ 
B_i^{in^{\prime}}, B_i^{out^{\prime}} \}$, EFCR$_{ex}$ with $\{ \mathcal{L}_i^{+^{\prime}},\mathcal{L}_i^{-^{\prime}}, \mathcal{L}_i^{ext^{\prime}} \}$ can be formulated as:
\begin{align}
      & \mathcal{L}_i^{+^{\prime}}= {\parallel}f^h(I_i^{out^{\prime}})-f^h(I_i^{gt^{\prime}}){\parallel}_1, \label{eq:posCR} \\
      & \mathcal{L}_i^{-^{\prime}}= {\parallel}f^h(I_i^{out^{\prime}})-f^h(I_i^{in^{\prime}}){\parallel}_1, \label{eq:negCR} \\
      & \mathcal{L}_i^{ext^{\prime}}= \frac{{\parallel}f^h(B_i^{out^{\prime}})-f^h(B_i^{in^{\prime}}){\parallel}_1}{{\parallel}f^h(I_i^{in^{\prime}})-f^h(B_i^{in^{\prime}}){\parallel}_1}.
  \vspace{-2mm}
\end{align}
The overall optimization objective follows a similar pattern with \cref{eq:overallCR}, where the supervised training on the testing dataset ($\mathcal{L}_1$) proceeds simultaneously with EFCR$_{ex}$ ($\mathcal{L}_{CR}$).
% to transfer knowledge without interfering with supervised training.

EFCR$_{ex}$ facilitates two important applications. 
One is to transfer rich deblur signals from a real-world blur dataset to microscopy deblur tasks, in which EFCR$_{ex}$ is composed by $\{ \mathcal{L}_i^{+^{\prime}},\mathcal{L}_i^{-^{\prime}}, \mathcal{L}_i^{ext^{\prime}} \}$.
Another is to learn latent deblur direction from an unlabeled microscopy dataset thus enhancing the deblur performance, where the model is trained on a labeled dataset with an unlabeled microscopy dataset as the extra data. EFCR$_{ex}$ here is reformulated as $\{ \mathcal{L}_i^{+},\mathcal{L}_i^{-}, \mathcal{L}_i^{ext^{\prime}} \}$.
% Another is to enhance the deblur performance on unsupervised domain of unlabeled microscopy images, which addresses the challenge of the lack of ground truth due to the inherent limitations of the microscopy as discussed in  \cref{sec:intro}. 
% In this scenario, the model is trained with a real-world dataset with unlabeled blurry microscopy images as extra data to learn the latent unlabeled deblur information, with EFCR $\{ \mathcal{L}_i^{+},\mathcal{L}_i^{-}, \mathcal{L}_i^{ext^{\prime}} \}$ where $\mathcal{L}_i^{+}$ and $\mathcal{L}_i^{-}$ involves real-world dataset and $\mathcal{L}_i^{ext^{\prime}}$ uses extra unlabeled blurry microscopy image.

\section{Experiments}
\label{sec:experiments}
%-------------------------------------------------------------------------

\begin{table*}
  \centering
  \setlength{\tabcolsep}{1.1mm}{} 
  \scalebox{0.95}{
  \begin{tabular}{lc@{}lc@{}lc@{}lc@{}lc@{}lc@{}lc@{}lc@{}}
    \toprule
\multirow{2}{*}{Method} & \multicolumn{3}{c}{BBBC006$_{w1}$ \cite{ljosa_annotated_2012}} & \multicolumn{3}{c}{BBBC006$_{w2}$ \cite{ljosa_annotated_2012}} & \multicolumn{3}{c}{3DHistech \cite{geng_cervical_2022}} & \multicolumn{3}{c}{CaDISBlur} \\
    \cmidrule{2-13}
& \multicolumn{1}{c}{PSNR$\uparrow$} & \multicolumn{1}{c}{SSIM$\uparrow$} & \multicolumn{1}{c}{LPIPS$\downarrow$} & \multicolumn{1}{c}{PSNR$\uparrow$} & \multicolumn{1}{c}{SSIM$\uparrow$} & \multicolumn{1}{c}{LPIPS$\downarrow$} & \multicolumn{1}{c}{PSNR$\uparrow$} & \multicolumn{1}{c}{SSIM$\uparrow$} & \multicolumn{1}{c}{LPIPS$\downarrow$} & \multicolumn{1}{c}{PSNR$\uparrow$} & \multicolumn{1}{c}{SSIM}$\uparrow$ & \multicolumn{1}{c}{LPIPS$\downarrow$}\\
    \midrule
DRBNet \cite{ruan_learning_2022} & \multicolumn{1}{c}{32.83} & \multicolumn{1}{c}{0.737} & \multicolumn{1}{c}{0.381} & \multicolumn{1}{c}{26.66} & \multicolumn{1}{c}{0.589} & \multicolumn{1}{c}{0.458} & \multicolumn{1}{c}{32.83} & \multicolumn{1}{c}{0.853} & \multicolumn{1}{c}{0.131} & \multicolumn{1}{c}{42.54} & \multicolumn{1}{c}{0.776} & \multicolumn{1}{c}{0.243} \\
GKMNet \cite{quan_gaussian_2021} & \multicolumn{1}{c}{34.41} & \multicolumn{1}{c}{0.887} & \multicolumn{1}{c}{0.218} & \multicolumn{1}{c}{29.32} & \multicolumn{1}{c}{0.721} & \multicolumn{1}{c}{0.296} & \multicolumn{1}{c}{33.42} & \multicolumn{1}{c}{0.852} & \multicolumn{1}{c}{0.130} & \multicolumn{1}{c}{44.27} & \multicolumn{1}{c}{0.860} & \multicolumn{1}{c}{0.178} \\
% DeblurGANv2 & \multicolumn{1}{c}{PSNR} & \multicolumn{1}{c}{SSIM} & \multicolumn{1}{c}{LPIPS} & \multicolumn{1}{c}{PSNR} & \multicolumn{1}{c}{SSIM} & \multicolumn{1}{c}{LPIPS} & \multicolumn{1}{c}{PSNR} & \multicolumn{1}{c}{SSIM} & \multicolumn{1}{c}{LPIPS} & \multicolumn{1}{c}{PSNR} & \multicolumn{1}{c}{SSIM} & \multicolumn{1}{c}{LPIPS} \\
MIMO-UNet \cite{cho_rethinking_2021} & \multicolumn{1}{c}{32.73} & \multicolumn{1}{c}{0.725} & \multicolumn{1}{c}{0.412} & \multicolumn{1}{c}{26.90} & \multicolumn{1}{c}{0.601} & \multicolumn{1}{c}{0.457} & \multicolumn{1}{c}{32.40} & \multicolumn{1}{c}{0.837} & \multicolumn{1}{c}{0.169} & \multicolumn{1}{c}{43.36} & \multicolumn{1}{c}{0.823} & \multicolumn{1}{c}{0.197} \\
MSSNet \cite{kim_mssnet_2023} & \multicolumn{1}{c}{34.01} & \multicolumn{1}{c}{0.790} & \multicolumn{1}{c}{0.289} & \multicolumn{1}{c}{28.68} & \multicolumn{1}{c}{0.736} & \multicolumn{1}{c}{0.361} & \multicolumn{1}{c}{33.09} & \multicolumn{1}{c}{0.870} & \multicolumn{1}{c}{0.126} & \multicolumn{1}{c}{44.09} & \multicolumn{1}{c}{0.871} & \multicolumn{1}{c}{0.160} \\
% MSSNet$_L$ & \multicolumn{1}{c}{PSNR} & \multicolumn{1}{c}{SSIM} & \multicolumn{1}{c}{LPIPS} & \multicolumn{1}{c}{PSNR} & \multicolumn{1}{c}{SSIM} & \multicolumn{1}{c}{LPIPS} & \multicolumn{1}{c}{PSNR} & \multicolumn{1}{c}{SSIM} & \multicolumn{1}{c}{LPIPS} & \multicolumn{1}{c}{PSNR} & \multicolumn{1}{c}{SSIM} & \multicolumn{1}{c}{LPIPS} \\
SwinIR \cite{liang_swinir_2021} & \multicolumn{1}{c}{33.90} & \multicolumn{1}{c}{0.801} & \multicolumn{1}{c}{0.274} & \multicolumn{1}{c}{27.61} & \multicolumn{1}{c}{0.696} & \multicolumn{1}{c}{0.403} & \multicolumn{1}{c}{32.57} & \multicolumn{1}{c}{0.841} & \multicolumn{1}{c}{0.136} & \multicolumn{1}{c}{41.83} & \multicolumn{1}{c}{0.710} & \multicolumn{1}{c}{0.349} \\
PANet \cite{mei_pyramid_2023} & \multicolumn{1}{c}{34.45} & \multicolumn{1}{c}{0.890} & \multicolumn{1}{c}{0.230} & \multicolumn{1}{c}{29.07} & \multicolumn{1}{c}{0.743} & \multicolumn{1}{c}{0.290} & \multicolumn{1}{c}{33.24} & \multicolumn{1}{c}{0.869} & \multicolumn{1}{c}{0.129} & \multicolumn{1}{c}{44.49} & \multicolumn{1}{c}{0.917} & \multicolumn{1}{c}{0.134} \\
GRL \cite{li_efficient_2023} & \multicolumn{1}{c}{34.76} & \multicolumn{1}{c}{{\color{blue}0.907}} & \multicolumn{1}{c}{{\color{blue}0.129}} & \multicolumn{1}{c}{29.39} & \multicolumn{1}{c}{0.786} & \multicolumn{1}{c}{0.249} & \multicolumn{1}{c}{{\color{blue}33.49}} & \multicolumn{1}{c}{0.878} & \multicolumn{1}{c}{{\color{red}0.120}} & \multicolumn{1}{c}{{\color{blue}44.86}} & \multicolumn{1}{c}{{\color{blue}0.960}} & \multicolumn{1}{c}{{\color{red}0.087}} \\
Restormer \cite{zamir_restormer_2022} & \multicolumn{1}{c}{{\color{blue}34.79}} & \multicolumn{1}{c}{0.904} & \multicolumn{1}{c}{0.135} & \multicolumn{1}{c}{{\color{blue}29.78}} & \multicolumn{1}{c}{{\color{blue}0.801}} & \multicolumn{1}{c}{{\color{blue}0.241}} & \multicolumn{1}{c}{33.46} & \multicolumn{1}{c}{{\color{blue}0.880}} & \multicolumn{1}{c}{0.125} & \multicolumn{1}{c}{44.85} & \multicolumn{1}{c}{0.941} & \multicolumn{1}{c}{0.101} \\
    \cmidrule{1-13}
% Uformer & \multicolumn{1}{c}{PSNR} & \multicolumn{1}{c}{SSIM} & \multicolumn{1}{c}{LPIPS} & \multicolumn{1}{c}{PSNR} & \multicolumn{1}{c}{SSIM} & \multicolumn{1}{c}{LPIPS} & \multicolumn{1}{c}{PSNR} & \multicolumn{1}{c}{SSIM} & \multicolumn{1}{c}{LPIPS} & \multicolumn{1}{c}{PSNR} & \multicolumn{1}{c}{SSIM} & \multicolumn{1}{c}{LPIPS} \\
MPT & \multicolumn{1}{c}{{\color{red}34.89}} & \multicolumn{1}{c}{{\color{red}0.912}} & \multicolumn{1}{c}{{\color{red}0.127}} & \multicolumn{1}{c}{{\color{red}29.85}} & \multicolumn{1}{c}{{\color{red}0.813}} & \multicolumn{1}{c}{{\color{red}0.237}} & \multicolumn{1}{c}{{\color{red}33.55}} & \multicolumn{1}{c}{{\color{red}0.881}} & \multicolumn{1}{c}{{\color{blue}0.121}} & \multicolumn{1}{c}{{\color{red}44.98}} & \multicolumn{1}{c}{{\color{red}0.962}} & \multicolumn{1}{c}{{\color{red}0.087}} \\
{MPT+EFCR} & \multicolumn{1}{c}{\underline{34.96}} & \multicolumn{1}{c}{\underline{0.917}} & \multicolumn{1}{c}{\underline{0.119}} & \multicolumn{1}{c}{\underline{29.89}} & \multicolumn{1}{c}{\underline{0.820}} & \multicolumn{1}{c}{\underline{0.230}} & \multicolumn{1}{c}{\underline{33.58}} & \multicolumn{1}{c}{\underline{0.887}} & \multicolumn{1}{c}{\underline{0.119}} & \multicolumn{1}{c}{\underline{45.09}} & \multicolumn{1}{c}{\underline{0.969}} & \multicolumn{1}{c}{\underline{0.082}} \\
{MPT+EFCR$_{ex}$} & \multicolumn{1}{c}{\underline{35.16}} & \multicolumn{1}{c}{\underline{0.935}} & \multicolumn{1}{c}{\underline{0.083}} & \multicolumn{1}{c}{\underline{30.11}} & \multicolumn{1}{c}{\underline{0.829}} & \multicolumn{1}{c}{\underline{0.205}} & \multicolumn{1}{c}{\underline{33.63}} & \multicolumn{1}{c}{\underline{0.892}} & \multicolumn{1}{c}{\underline{0.116}} & \multicolumn{1}{c}{\underline{45.25}} & \multicolumn{1}{c}{\underline{0.971}} & \multicolumn{1}{c}{\underline{0.077}} \\
    \bottomrule
  \end{tabular}
  }
  \caption{Quantitative evaluation on microscopy deblur. The experiments on the sub-set $w1$ (stained by Hoechst to show nuclei structure) and $w2$ (stained by phalloidin to show cell structure) of BBBC006 \cite{ljosa_annotated_2012} are conducted separately. Except for the methods using \underline{EFCR}, the methods with the best and second best performance are noted in {\color{red}red} and {\color{blue}blue} colors.}
  \label{tab:resultmicro}
\end{table*}

\begin{figure*}[t]
  \centering
   \includegraphics[width=0.93\linewidth]{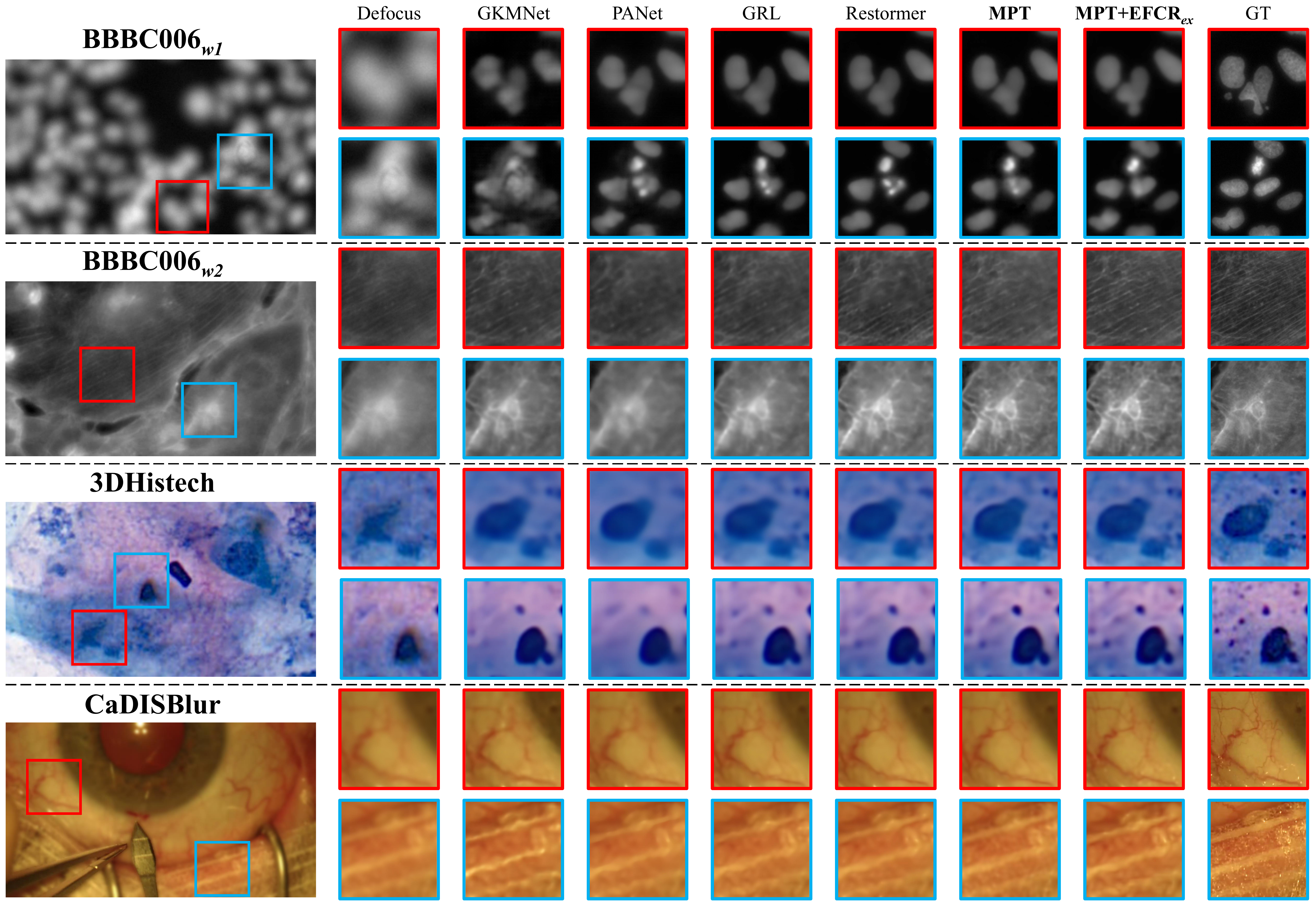}
   % \vspace{-2mm}
   \caption{Qualitative evaluation on microscopy deblur. Our method achieves the best restoration of different types of defocus blur. \vspace{-3mm}}
   \label{fig:qualevallabeled}
\end{figure*}

\begin{table}
  \centering
  \setlength{\tabcolsep}{3.5mm}{} 
  \scalebox{1.0}{
  \begin{tabular}{@{}lc@{}lc@{}lc@{}lc}
    \toprule
% \multirow{2}{*}{Method} & \multicolumn{3}{c}{DPDD} & \multicolumn{3}{c}{LFDOF} \\
%     \cmidrule{2-7}
Method & \multicolumn{1}{c}{PSNR$\uparrow$} & \multicolumn{1}{c}{SSIM$\uparrow$} & \multicolumn{1}{c}{LPIPS$\downarrow$}\\
    \midrule
% DPDNet$_S$ & \multicolumn{1}{c}{24.39} & \multicolumn{1}{c}{0.749} & \multicolumn{1}{c}{0.277}\\
KPAC \cite{son_single_2021} & \multicolumn{1}{c}{25.24} & \multicolumn{1}{c}{0.774} & \multicolumn{1}{c}{0.226}\\ 
IFAN \cite{lee_iterative_2021} & \multicolumn{1}{c}{25.37} & \multicolumn{1}{c}{0.789} & \multicolumn{1}{c}{0.217}\\
% AIFNet & \multicolumn{1}{c}{24.21} & \multicolumn{1}{c}{0.742} & \multicolumn{1}{c}{0.309} \\
DRBNet \cite{ruan_learning_2022} & \multicolumn{1}{c}{25.47} & \multicolumn{1}{c}{0.787} & \multicolumn{1}{c}{0.246} \\
% DRBNet$_P$ & \multicolumn{1}{c}{25.73} & \multicolumn{1}{c}{0.791} & \multicolumn{1}{c}{0.183}\\
GKMNet \cite{quan_gaussian_2021} & \multicolumn{1}{c}{25.47} & \multicolumn{1}{c}{0.789} & \multicolumn{1}{c}{0.219} \\
Restormer \cite{zamir_restormer_2022} & \multicolumn{1}{c}{25.98} & \multicolumn{1}{c}{0.811} & \multicolumn{1}{c}{0.178} \\
NRKNet \cite{quan_neumann_2023} & \multicolumn{1}{c}{26.11} & \multicolumn{1}{c}{0.810} & \multicolumn{1}{c}{0.210} \\
GRL \cite{li_efficient_2023} & \multicolumn{1}{c}{{\color{blue}26.18}} & \multicolumn{1}{c}{{\color{blue}0.822}} & \multicolumn{1}{c}{{\color{red}0.168}} \\
\midrule
MPT & \multicolumn{1}{c}{{\color{red}26.21}} & \multicolumn{1}{c}{{\color{red}0.826}} & \multicolumn{1}{c}{{\color{blue}0.175}}\\
MPT+EFCR & \multicolumn{1}{c}{\underline{26.23}} & \multicolumn{1}{c}{\underline{0.829}} & \multicolumn{1}{c}{\underline{0.172}}\\
MPT+EFCR$_{ex}$ & \multicolumn{1}{c}{\underline{26.27}} & \multicolumn{1}{c}{\underline{0.831}} & \multicolumn{1}{c}{\underline{0.161}}\\
    \bottomrule
  \end{tabular}
  }
  \caption{Quantitative real-world deblur evaluation on DPDD \cite{abuolaim_defocus_2020}.\vspace{-2mm}}
  \label{tab:resultDPDD}
% \vspace{-2mm}
\end{table}

\subsection{Datasets and Implementation}
\label{sec:experiments4.1}
Extensive experiments are carried out on various real-world and microscopy datasets, including five labeled datasets: DPDD \cite{abuolaim_defocus_2020}, LFDOF \cite{ruan_aifnet_2021}, BBBC006 \cite{ljosa_annotated_2012}, 3DHistech \cite{geng_cervical_2022}, CaDISBlur, and three unlabeled datasets: CUHK \cite{shi_discriminative_2014}, WNLO \cite{geng_cervical_2022}, CataBlur.
The LFDOF \cite{ruan_aifnet_2021} is adopted as an extra training dataset for knowledge transfer using EFCR, since LFDOF has substantial samples with rich information and good cross correlation between defocused and ground truth pairs \cite{ruan_learning_2022}.
% For BBBC006 \cite{ljosa_annotated_2012}, the test is carried out on two sub-sets respectively, which are noted as \textit{w1} stained by Hoechst for nuclei structure and \textit{w2} stained by phalloidin for cell structure.
For surgical microscopy, two new surgical microscopy deblur datasets are presented, which are CaDISBlur and CataBlur.
CaDISBlur is synthesized using images from a high-quality dataset CaDIS \cite{grammatikopoulou_cadis_2022} by a novel realistic blur simulation method, in which the instruments and anatomies in the surgery scene are blurred respectively leveraging the object segmentation mask in CaDIS to simulate different focal planes.
% instead of uniformly blurring the whole image \cite{ghamsarian_deblurring_2020}.
CataBlur is a new surgical microscope defocus blur dataset, including 1208 defocus images collected from 5 cataract surgeries, for evaluation on real surgical defocus blur.
More details about the datasets, training settings, and proposed blur synthesizing method are provided in the supplementary material.

The proposed framework employs the same structure in all tests as follows.
The MPT adopts a 4-stage design as shown in \cref{fig:MPT}, with $[6,6,6,6]$ sub-blocks, $[40,80,160,320]$ feature dimensions, and $[1,2,4,8]$ attention heads. 
The scale set of each pyramid block is set as $S_1=[\frac{1}{8},\frac{1}{8},\frac{1}{4},\frac{1}{4},1,1]$, $S_2=[\frac{1}{4},\frac{1}{4},\frac{1}{2},\frac{1}{2},1,1]$, $S_3=S_4=[\frac{1}{2},\frac{1}{2},\frac{1}{2},\frac{1}{2},1,1]$.
The expansion ratio $\alpha$ in FEFN is set to 2.6, and the scaling factor $\beta$ in EFCR is set to 1e$^{-5}$.
The method is implemented using PyTorch and trained with AdamW optimizer \cite{loshchilov2017decoupled} ($\beta_1=0.9$, $\beta_2=0.999$, weight decay is 1e$^{-4}$) for $3\times 10^5$ iterations on NVIDIA A800 GPUs. 
The initial learning rate is set to 1e$^{-4}$ and gradually decreases to 1e$^{-6}$ by cosine annealing \cite{loshchilov2016sgdr}. 
The batch size is set to 8 with training patches in the size of 256$\times$256 augmented with random scaling, and horizontal and vertical flips.
% Two implementations of our framework is included for comparison. One is the MPT and EFCR without extra training data, noted as \textbf{Ours}. The other one is the MPT and EFCR with LFDOF \cite{ruan_aifnet_2021} as extra data trained by the paradigm described in \cref{sec:methodEFCR}, noted as \textbf{Ours$_{\textbf{ex}}$}.
Three implementations are included, which are MPT, MPT with EFCR, and MPT with EFCR using LFDOF as extra data (noted as EFCR$_{ex}$).
Then the result is reported in three metrics, including Peak Signal-to-Noise Ratio (PSNR), Structural Similarity (SSIM) \cite{wang2004image}, and Learned Perceptual Image Patch Similarity (LPIPS) \cite{zhang2018unreasonable}.

\subsection{Comparison and Analysis}
\paragraph{Evaluation on supervised deblur}
The evaluation of cell microscopy deblur and surgical microscopy deblur is conducted on three microscopy datasets covering a wide range of state-of-the-art defocus deblur methods and image restoration methods. 
Real-world deblur evaluation is also conducted on DPDD \cite{abuolaim_defocus_2020} to further prove the generalizability and universality.
The result is shown in \cref{tab:resultmicro} and \cref{tab:resultDPDD}.
The proposed framework demonstrates satisfactory performance on all microscopy datasets and real-world datasets, showing the advantages of the proposed MPT structure and EFCR training strategy.
% It is worth noting that our method achieves best performance while keeping computational efficiency. 
Compared with Restormer \cite{zamir_restormer_2022}, which achieves the second-best performance on BBBC006 \cite{ljosa_annotated_2012}, MPT (76 FLOPs, 19.80 M) outperforms Restormer (141 FLOPs, 26.12 M) by 0.10 dB and 0.07 dB regarding PSNR while saving 46\% FLOPs (for a 256$\times$256 input) and 24\% parameters, since MPT extracts richer representation than Restormer which only applies channel attention.
GRL \cite{li_efficient_2023} achieves the second-best performance on 3DHistech \cite{geng_cervical_2022} and CaDISBlur, but it directly models global spatial attention without leveraging the properties of downscaled maps like CSWA, resulting in 1230 FLOPs for a 256$\times$256 input that is 17$\times$ larger than ours.
Compared to MSSNet \cite{kim_mssnet_2023}, MIMO-UNet \cite{cho_rethinking_2021}, and PANet \cite{mei_pyramid_2023} that adopt multi-scale or pyramid design, our method with multi-pyramid structure outperforms them in all tests.
SwinIR \cite{liang_swinir_2021} adopts the original local window attention \cite{liu_swin_2021}, yet is hindered by the limited receptive field and fails to build long-range interaction.
The visualizations shown in \cref{fig:qualevallabeled} prove that our method achieves the best restoration of fine details against strong defocus blur, especially for the miniature cell shape and complex cell structure, as well as precise features of surgical anatomies. 
For real-world deblur on DPDD \cite{abuolaim_defocus_2020}, our method achieves the best performance in terms of SSIM and PSNR.
% Although the LPIPS is slightly lower than the second-best method, it still surpasses the second-best method NRKNet \cite{quan_neumann_2023} by 17\%.
It shows that our model is universally applicable to different types of images.
Visualization of deblurring on DPDD is provided in \cref{fig:sup_DPDD} in supplementary materials.

% which is critical for microscopy tasks as analyzed in \cref{sec:intro}.
% Our MPT outperforms SwinIR \cite{liang_swinir_2021} by leveraging enlarged receptive field in CSWA, without introducing addition complexity. 
% Our model outperforms the methods that adopts window attention (SwinIR \cite{liang_swinir_2021}) and channel attention (Restormer \cite{zamir_restormer_2022})
% It also indicates better robustness and generalizability of the proposed method than the existing methods, since the previous models fail to achieve consistent excellency across various image and blur types, such as XXX, which achieve XXX in XXX but only XXX in XXX.

For MPT trained with EFCR, the performance is improved by learning latent deblur information.
% Noted that the EFCR training strategy is also applicable to the other methods as shown in \cref{sec:discussion}.
By further applying EFCR$_{ex}$ to learn cross-domain deblur guidance, the deblur performance is significantly enhanced in all four microscopy datasets.
It proves that deblurring benefits from cross-domain knowledge, despite the significant feature discrepancy between real-world extra data and microscope images.
Improvements are also observed in SSIM and LPIPS, showing that EFCR$_{ex}$ enhances deblurring from the perspective of the human visual system, which is of great significance for clinical application.
Visualizations in \cref{fig:qualevallabeled} draw a similar conclusion that the model with EFCR$_{ex}$ can restore the fine details more precisely.
Real-world deblur can also benefit from EFCR as all three metrics are improved by integrating EFCR or EFCR$_{ex}$.
% By further applying EFCR$_{ex}$, significant enhancement is observed, showing the knowledge transfer with EFCR is effective for natural scene.
Further discussion in \cref{sec:discussion} shows the superiority of this proposed training diagram against simply pretraining and fine-tuning.

\begin{figure}[t]
  \centering
   \includegraphics[width=1\linewidth]{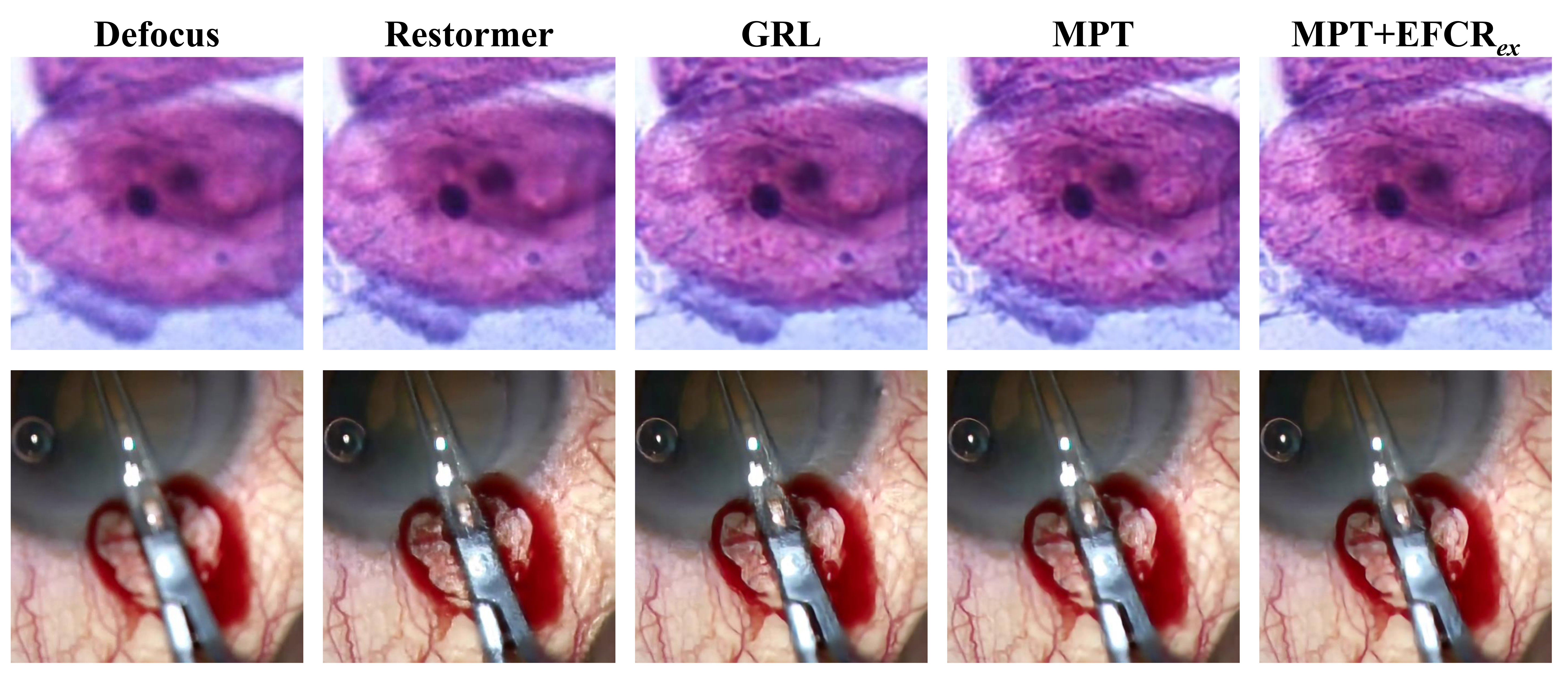}
   \caption{Qualitative evaluation on unsupervised deblur with WNLO (top) and CataBlur (bottom).}
    % {\color{blue}I actually cannot see the differences or the superiority of MPT and MPT+EFCR.}
   \label{fig:qualevalunlabeled}
\end{figure}

% \vspace{-3mm}
\paragraph{Evaluation on unsupervised deblur}
The deblur experiments on unlabeled datasets are conducted to qualitatively evaluate the generalizability of the model, and also to prove the effectiveness of knowledge transfer based on the proposed EFCR$_{ex}$.
Two unlabeled microscopy blur datasets are involved, including WNLO \cite{geng_cervical_2022} and CataBlur.  
% Four methods are taken for comparison, which are Restormer \cite{zamir_restormer_2022}, GRL \cite{li_efficient_2023}, MSSNet \cite{kim_mssnet_2023}, and GKMNet \cite{quan_gaussian_2021}.
All methods are trained on LFDOF since the rich information in the real-world dataset benefits microscopy deblur (see proof in supplementary materials).
Unlabeled datasets are adopted as the extra data to learn cross-domain latent information using EFCR$_{ex}$.
The qualitative comparison is shown in \cref{fig:qualevalunlabeled}.
Even without EFCR$_{ex}$, our method still shows the best generalizability with fewer artifacts and successfully restores the detail from strong defocus degradation.
With the help of EFCR$_{ex}$, the artifacts are further reduced, resulting in clearer deblurred images.
Results on the real-world dataset CUHK \cite{shi_discriminative_2014} are shown in \cref{fig:sup_CUHK} in supplementary materials, which demonstrates the universality of our method.

\begin{table*}[!h]
\floatsetup{floatrowsep=qquad,captionskip=10pt} \tabcolsep=8pt 
\begin{floatrow}

\begin{minipage}[l]{0.3\textwidth}
\ttabbox{\caption{Ablation studies on attention blocks with four variants, where WA refers to the original version of window attention \cite{liang_swinir_2021,liu_swin_2021}. Performance degradation occurs in all variants.}}
{
\scalebox{0.77}{
\setlength{\tabcolsep}{0.4mm}{} 
  \begin{tabular}{lccc}
    \toprule
 \multicolumn{1}{l}{Configuration}  & \multicolumn{1}{c}{PSNR$_D$} & PSNR$_B$& PSNR$_C$\\
    \midrule
$V_1$ (CSWA$\times$2)  & \multicolumn{1}{c}{26.10} & 34.76& 44.83\\ % 76 FLOPs 19.80 M
$V_2$ (WA$\times$2)& 25.92& 33.98&44.02\\ % 81 FLOPs 19.12 M
$V_3$ (ISCA$\times$2) & \multicolumn{1}{c}{26.01} & 34.60& 44.71\\ % 
$V_4$ (WA+ISCA) & 26.13& 34.10&44.78\\ % 81 FLOPs 19.12 M
MPT (CSWA+ISCA)& 26.21& 34.89& 44.98\\ % 76 FLOPs 19.80 M
    \bottomrule
\label{tab:ablationcswa}
  \end{tabular}
  }}
% \captionof{table}{Ablation studies 1.}
\end{minipage}
\hspace{1mm}
\begin{minipage}[c]{0.3\textwidth}
\ttabbox{\caption{Ablation studies on FEFN with three variants, which are symmetric structures: concatenation ($V_1$) and adding ($V_2$) followed by GELU activation, and reversed structure that uses the feature from CSWA for activation ($V_3$)}}{
\vspace{-0.98mm}
\scalebox{0.77}{
\setlength{\tabcolsep}{0.4mm}{} 
  \begin{tabular}{lccc}
    \toprule
 \multicolumn{1}{l}{Configuration}  & \multicolumn{1}{c}{PSNR$_D$} & PSNR$_B$& PSNR$_C$\\
    \midrule
$V_1$ (Cat+GELU)& \multicolumn{1}{c}{26.18} & 34.86& 44.84\\  % 87 FLOPs 22.45 M
$V_2$ (Add+GELU)& 26.03& 34.75&44.87\\
 $V_3$ (reversed)& 25.98& 34.72&44.50\\  % 76 FLOPs 19.80 M
  MPT (FEFN)& 26.21& 34.89& 44.98\\  % 76 FLOPs 19.80 M
    \bottomrule
\label{tab:ablationfefn}
  \end{tabular}
  }}
% \captionof{table}{Ablation studies 1.}
\end{minipage}
\hspace{1mm}
\begin{minipage}[c]{0.3\textwidth} 
\ttabbox{\caption{Ablation studies on EFCR. $\Delta$PSNR refers to the changes in PSNR compared to the baseline.}}{ \scalebox{0.77}{
\setlength{\tabcolsep}{0.1mm}{}
  \begin{tabular}{lccc}
    \toprule
 \multicolumn{1}{l}{Configuration}  & \multicolumn{1}{c}{$\Delta$PSNR$_D$} & $\Delta$PSNR$_B$& $\Delta$PSNR$_C$\\
    \midrule
MPT+$V_1$ & \multicolumn{1}{c}{+0.01} & +0.04& +0.05\\
 MPT+EFCR& +0.02& +0.07&+0.11\\
 Restormer+EFCR& +0.04& +0.07&+0.09\\
    \midrule
 MPT+pretrain& +0.05& -0.02&+0.01\\
 MPT+$V_{ex1}$& +0.05& +0.21&+0.18\\
 MPT+EFCR$_{ex}$& +0.06& +0.27&+0.27\\
 Restormer+EFCR$_{ex}$& +0.07& +0.19&+0.21\\
     \bottomrule
  \label{tab:ablationefcr}
  \end{tabular}
  }}
% \captionof{table}{Ablation studies 1.}
\end{minipage}
\vspace{-6mm}
\end{floatrow}
\end{table*}

\subsection{Validation on downstream tasks}
To demonstrate the clinical-related improvement, validations on medical downstream tasks are conducted.
% To further show the superior deblur performance of the proposed framework, two microscopy scenarios suffering from defocus blur are evaluated, including cell detection on BBBC006 \cite{ljosa_annotated_2012}, and surgical scene semantic segmentation on CaDIS \cite{grammatikopoulou_cadis_2022}.

\begin{figure}[!t]
  \centering
   \includegraphics[width=1.0\linewidth]{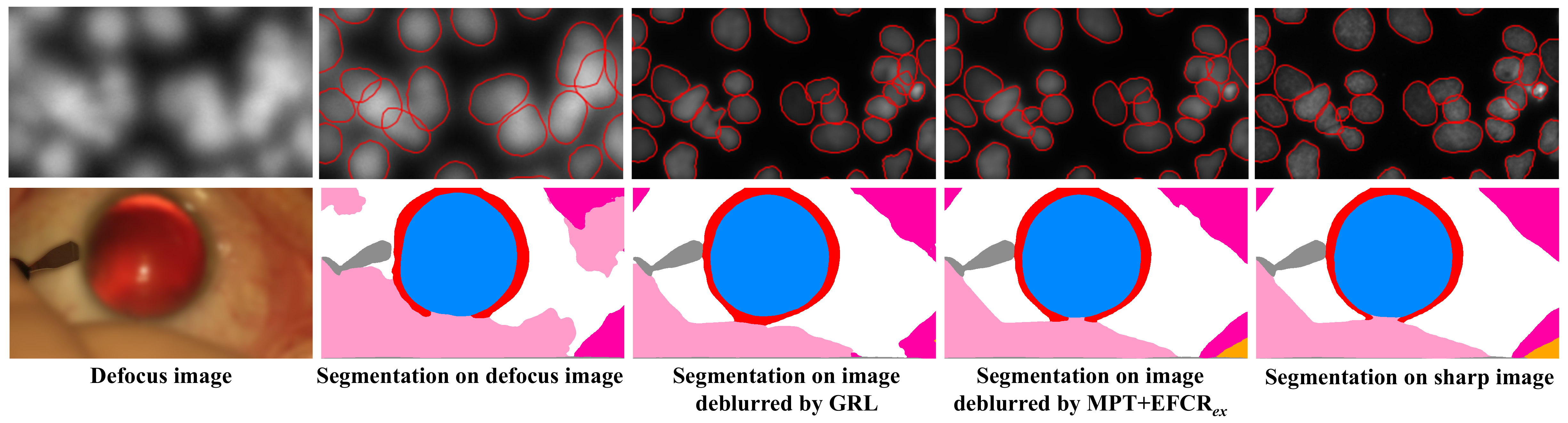}
   \caption{Downstream tasks result on BBBC006 (top) and CaDISBlur (bottom). Our method leads to less false segmentation. \vspace{-4mm}}
   % {\color{blue}I hope this figure can be larger if possible.}
   \label{fig:downstream}
\end{figure}

\paragraph{Cell detection on BBBC006}
Defocus blur can cause failure in cell detection and segmentation \cite{geng_cervical_2022,chen_automatic_2022} that is essential for many biological tasks \cite{meijering2012cell}.
The cell segmentation is performed using StarDist  \cite{schmidt_cell_2018}  on images in BBBC006 before and after deblur.
The result is reported in \cref{tab:resultBBBC} regarding the average precision (AP) over different intersection-of-union (IoU) thresholds, where higher AP means more cells are successfully detected.
Our deblur framework significantly improves the cell detection performance by 19.57\% (with extra data) and 18.54\% (without extra data) compared with blurry input, surpassing the improvement brought by Restormer (16.05\%) and GRL (13.50\%).
% A remarkable result in AP over a 0.9 IoU threshold, surpassing the other two methods by 17.90\% and 28.63\% even without extra data, demonstrating its superior accuracy in restoring the cell shape. 
The visualization shown in \cref{fig:downstream} also proves that our method achieves better restoration of the cell shape and structure.

\begin{table}
  \centering
  \setlength{\tabcolsep}{1.7mm}{} 
  \scalebox{0.9}{
  \begin{tabular}{lcccc}
    \toprule
IoU& \multicolumn{1}{c}{0.5} & \multicolumn{1}{c}{0.7} & \multicolumn{1}{c}{0.9} & Mean AP\\
    \midrule
 blur& 0.7010& 0.5623& 0.2194& 0.4942\\
    \midrule
Restormer \cite{zamir_restormer_2022} & \multicolumn{1}{c}{0.7789} & \multicolumn{1}{c}{0.6703} & \multicolumn{1}{c}{0.2714} & 0.5735 \\
GRL \cite{li_efficient_2023}& \multicolumn{1}{c}{0.7702} & \multicolumn{1}{c}{0.6433} & \multicolumn{1}{c}{0.2691} & 0.5609 \\
GRL+EFCR & \multicolumn{1}{c}{0.7710} & \multicolumn{1}{c}{0.6440} & \multicolumn{1}{c}{0.2695} & 0.5615 \\
GRL+EFCR$_{ex}$ & \multicolumn{1}{c}{0.7769} & \multicolumn{1}{c}{0.6491} & \multicolumn{1}{c}{0.2733} & 0.5664 \\
MPT (w/o EFCR) & \multicolumn{1}{c}{0.7808} & \multicolumn{1}{c}{0.6778} & \multicolumn{1}{c}{0.2956} &0.5847\\
MPT+EFCR & \multicolumn{1}{c}{{\color{blue}0.7814}} & \multicolumn{1}{c}{{\color{blue}0.6791}} & \multicolumn{1}{c}{{\color{blue}0.2970}} &{\color{blue}0.5858}\\
MPT+EFCR$_{ex}$ & \multicolumn{1}{c}{{\color{red}0.7865}} & \multicolumn{1}{c}{{\color{red}0.6843}} & \multicolumn{1}{c}{{\color{red}0.3019}} & {\color{red}0.5909}\\
    \midrule
 sharp& 0.8021& 0.7192& 0.3518&0.6244\\
     \bottomrule
  \end{tabular}
  }
  \caption{\vspace{-2mm}Cell detection result on deblurred BBBC006.}
  \label{tab:resultBBBC}
\end{table}

% \vspace{-3mm}
\paragraph{Semantic segmentation on CaDISBlur}
Semantic segmentation plays an important role in surgical scene understanding \cite{grammatikopoulou_cadis_2022}.
The semantic segmentation of the cataract surgical scene is conducted using OCRNet \cite{yuan_object-contextual_2020} on CaDISBlur based on CaDIS \cite{grammatikopoulou_cadis_2022}.
Results of images with different focal plane positions (blurry instruments or blurry anatomies, noted as \textit{ins} or \textit{ana}) are reported separately in \cref{tab:resultcadisblur} regarding mean IoU (mIoU) and pixel accuracy (PA). 
% {\color{blue}Some analysis of the results should be provided here.}
The deblurred images from our method lead to the best performance in most metrics.
Visualizations shown in \cref{fig:downstream} also demonstrate the superiority of our method.

\begin{table}
  \centering
  \setlength{\tabcolsep}{1.3mm}{} 
  \scalebox{0.9}{
  \begin{tabular}{lcccc}
      \toprule
 \multirow{2}{*}{Method} & \multicolumn{2}{c}{Blurry instrument}& \multicolumn{2}{c}{Blurry anatomies}\\
     \cmidrule{2-5}
 & \multicolumn{1}{c}{mIoU$_{ins}$} & PA$_{ins}$& \multicolumn{1}{c}{mIoU$_{ana}$} & \multicolumn{1}{c}{PA$_{ana}$}\\
    \midrule
blur & \multicolumn{1}{c}{0.7577} & 0.8677& \multicolumn{1}{c}{0.7092} & \multicolumn{1}{c}{0.8194}\\
    \midrule
% Restormer \cite{zamir_restormer_2022} & \multicolumn{1}{c}{} & \multicolumn{1}{c}{} & \multicolumn{1}{c}{} \\
Restormer \cite{zamir_restormer_2022} & \multicolumn{1}{c}{0.7558} & 0.8803& \multicolumn{1}{c}{0.8149} & \multicolumn{1}{c}{0.8674} \\
GRL \cite{li_efficient_2023} & \multicolumn{1}{c}{0.7606} & {0.8849}& \multicolumn{1}{c}{0.8135} & \multicolumn{1}{c}{0.8689} \\
GRL+EFCR & \multicolumn{1}{c}{0.7611} & {0.8851}& \multicolumn{1}{c}{0.8140} & \multicolumn{1}{c}{0.8691} \\
GRL+EFCR$_{ex}$ & \multicolumn{1}{c}{\color{blue}0.7625} & {\color{blue}0.8857}& \multicolumn{1}{c}{0.8162} & \multicolumn{1}{c}{0.8710} \\
MPT (w/o EFCR) & \multicolumn{1}{c}{0.7607} & 0.8836& \multicolumn{1}{c}{0.8293} & \multicolumn{1}{c}{0.8824} \\
MPT+EFCR & \multicolumn{1}{c}{{0.7610}} & 0.8842& \multicolumn{1}{c}{{\color{blue}0.8295}} & \multicolumn{1}{c}{{\color{blue}0.8830}}\\
MPT+EFCR$_{ex}$ & \multicolumn{1}{c}{{\color{red}0.7667}} & {\color{red}0.8859}& \multicolumn{1}{c}{{\color{red}0.8361}} & \multicolumn{1}{c}{{\color{red}0.8896}}\\
    \midrule
sharp & \multicolumn{1}{c}{0.7733} & 0.8886& \multicolumn{1}{c}{0.8582} & \multicolumn{1}{c}{0.9305}\\
    \bottomrule
  \end{tabular}
  }
  \caption{\vspace{-2mm}Semantic segmentation result on deblurred CaDISBlur.}
  \label{tab:resultcadisblur}
\end{table}

\subsection{Ablation Studies}
\label{sec:discussion}
For ablation studies, the model variants are evaluated regarding PSNR on DPDD \cite{abuolaim_defocus_2020}, BBBC006$_{w1}$ \cite{ljosa_annotated_2012} and CaDISBlur datasets, which are denoted by PSNR$_D$, PSNR$_B$ and PSNR$_C$, respectively.
More ablation experiments and analyses are provided in supplementary materials.

\paragraph{Configurations of pyramid block}
% To explore the performance gain brought by CSWA and ISCA, experiments are conducted on four network variants, as shown in \cref{tab:ablationcswa}.
% As the result shows, performance degradation occurs in all variants.
Ablation studies on CSWA and ISCA are first carried out. As shown in \cref{tab:ablationcswa}, a performance drop is observed when changing CSWA to WA ($V_4$) since WA only models attention within a local window. Although hierarchical network structure in MPT may provide WA with a larger receptive field at low-resolution stages, it still causes inferior performance than CSWA since cross-scale interactions are not built.
The situation gets worse if the two attention blocks are all changed to WA ($V_2$) since the model lost long-range modeling ability in both channel and spatial means.
% Together with $V_1$ and $V_3$, degradation in $V_2$ also proves the necessity of adopting multi-type attention.
Experiments are then carried out on variants of FEFN, as shown in \cref{tab:ablationfefn}, which shows the superiority of the proposed asymmetrical activation.

\vspace{-3mm}
\paragraph{Improvements in EFCR}
% Two scenarios with or without involving extra data are evaluated separately.
The result is shown in \cref{tab:ablationefcr}.
$V_1$ and $V_{ex1}$ refer to $\{ \mathcal{L}_i^{+},\mathcal{L}_i^{-} \}$ and $\{ \mathcal{L}_i^{+^{\prime}},\mathcal{L}_i^{-^{\prime}} \}$. 
% without extended CR $\mathcal{L}_i^{ext}$ or $\mathcal{L}_i^{ext^{\prime}}$.
The proposed EFCR and EFCR$_{ex}$ yield significant improvements over baseline, not only on our method but also on Restormer \cite{zamir_restormer_2022}, proving the effectiveness of the proposed training diagram.
% $V_1$ and $V_{ex1}$ have less performance gain than EFCR and EFCR$_{ex}$, 
Following a similar approach in \cite{ruan_learning_2022} to pretrain and fine-tune (denoted as \textit{MPT+pretrain}), the trained model leads to a trivial improvement or even degradation.
It further demonstrates the superiority of our EFCR$_{ex}$.

% \begin{table}
%   \centering
%   \setlength{\tabcolsep}{1mm}{} 
%   \begin{tabular}{lccc}
%     \toprule
%  \multicolumn{1}{l}{Configuration}  & \multicolumn{1}{c}{$\Delta$PSNR$_D$} & $\Delta$PSNR$_B$& $\Delta$PSNR$_C$\\
%     \midrule
% MPT+$V_1$ & \multicolumn{1}{c}{+0.01} & +0.04& +0.05\\
%  MPT+EFCR& +0.02& +0.07&+0.11\\
%  Restormer+EFCR& +0.04& +0.07&+0.09\\
%     \midrule
%  MPT+pretrain& +0.05& -0.02&+0.01\\
%  MPT+$V_{ex1}$& +0.05& +0.21&+0.18\\
%  MPT+EFCR$_{ex}$& +0.06& +0.27&+0.27\\
%  Restormer+EFCR$_{ex}$& +0.07& +0.19&+0.21\\
%      \bottomrule
%   \end{tabular}
%   \caption{Ablation studies on EFCR. $\Delta$PSNR refers to the changes in PSNR compared to the baseline.}
%   \label{tab:ablationefcr}
% \end{table}

\vspace{-1mm}
\section{Conclusion}
\label{sec:conclusion}

%-------------------------------------------------------------------------

\vspace{-1mm}
This paper presents a unified framework to address the outstanding problems in microscopy defocus deblur.
The MPT outperforms existing multi-scale networks by incorporating spatial-channel context from CSWA and ISCA using FEFN. 
% the proposed network achieves consistent excellence across different datasets.  
% adapting to microscopy dataset with longer attention span
The proposed EFCR enforces the model to explore latent deblur guidance and further learn cross-domain knowledge from the extra data, yielding significant performance gain in both supervised and unsupervised image deblur. 
In the future, larger-scale datasets, e.g. ImageNet \cite{deng2009imagenet}, will be adopted for knowledge transfer using EFCR, along with experiments on weakly supervised or unsupervised learning \cite{he2024weakly} and domain adaptation \cite{zhang2022tdacnn}.
Experiments on MPT variants incorporating varied window mechanisms \cite{xia_vision_2022} will be carried out.

\vspace{-1mm}
\paragraph{Acknowledgement}
This work was supported in part by the Innovation and Technology Commission of Hong Kong under Grant ITS/233/21 and ITS/234/21, in part by the Multi-Scale Medical Robotics Center, InnoHK, and in part by CUHK Direct Grant. We would like to thank Dr. Danny Siu-Chun Ng from The Department of Ophthalmology and Visual Sciences at The Chinese University of Hong Kong for providing the cataract surgery images for research.

{
    \small
    \bibliographystyle{ieeenat_fullname}
    \bibliography{main,ref}
}

\clearpage
\setcounter{page}{1}
\maketitlesupplementary

\section{Dataset Description}
\label{sec:sup_dataset}

This section provides supplementary information about the datasets involved in the evaluation. The details are shown in \cref{tab:dataset}.
% As mentioned in \cref{sec:experiments}, experiments are conducted on a total of eight datasets, including three real-world datasets, three cell microscopy datasets, and two surgical microscopy datasets, as shown in the \ref{tab:dataset}.
% Extensive experiments are carried out on various real-world and microscopy datasets, including five labeled datasets: DPDD \cite{abuolaim_defocus_2020}, LFDOF \cite{ruan_aifnet_2021}, BBBC006 \cite{ljosa_annotated_2012}, 3DHistech \cite{geng_cervical_2022}, CaDISBlur, and three unlabeled datasets: CUHK \cite{shi_discriminative_2014}, WNLO \cite{geng_cervical_2022}, CataBlur.

For real-world scenarios, three datasets are included. For DPDD \cite{abuolaim_defocus_2020}, the sharp-blur training pairs are collected sequentially by a DSLR camera with different aperture sizes. 
% This procedure causes pixel incorrespondence between the sharp ground truth and blurry input, which is harmful to deblur model training \cite{ruan_learning_2022}. 
A labeled defocus deblur dataset LFDOF \cite{ruan_aifnet_2021} collected by a light field camera is adopted in this paper as extra data for knowledge transfer since the training pairs captured by the light field camera have strict pixel-wise consistency \cite{ruan_learning_2022}. 
LFDOF contains many more images than DPDD, which also benefits the microscopy deblur by transferring rich cross-domain information.
For unsupervised deblur, an unlabeled blur dataset CUHK \cite{shi_discriminative_2014} is adopted, which is collected from the internet.

Three datasets are involved in cell microscopy deblur. 
BBBC006 comes from the Broad Bioimage Benchmark Collection \cite{ljosa_annotated_2012}, which contains images in two sub-sets stained by Hoechst and phalloidin captured by fluorescence microscope.
It contains images with different focal planes (denoted as different z-stacks). 
Following dataset description \cite{ljosa_annotated_2012}, images collected on the optimal focal length (z-stack = 16) are set as the ground truth, and images above the optimal focal plane are used for training.
To avoid redundancy, images with z-stack = [2, 6, 10] are set as blurry input for training.
Since the images in BBBC006 only contain a single grayscale channel, for EFCR$_{ex}$ in BBBC006, the images in LFDOF are converted into grayscale with one channel.
3DHistech and WNLO \cite{geng_cervical_2022} are two cell imaging datasets for cytopathology scanned by digital scanners.
The labeled dataset 3DHistech is scanned using different focal planes, where the focal plane with the most cells in focus is set as the ground truth.
WNLO is an unlabeled dataset with defocus images only.
% TODO train test set setting

Regarding the surgical microscopy deblur, two new datasets are presented, which are the labeled synthesized dataset CaDISBlur and the unlabeled cataract surgery defocus blur dataset CataBlur.
% \Cref{fig:sup_cataexamples-a} and \Cref{fig:sup_cataexamples-b} shows some images in CaDISBlur and CataBlur.
CaDISBlur is synthesized based on CaDIS \cite{grammatikopoulou_cadis_2022}, which is a dataset for surgical scene semantic segmentation. Leveraging segmentation masks, the instruments and anatomies are blurred separately to simulate different focal planes. The original images in CaDIS are of high quality thus they can be treated as the sharp ground truth.
CataBlur is an unlabeled real defocus blur dataset containing 1208 images acquired during 5 different cataract surgeries, from which the severity of defocus blur in microscope surgery can be observed. 
The privacy information is removed. The experiment's conduction and the dataset's collection are granted with ethical approval.
The images in CataBlur are sampled from surgery videos with lower frames per second (fps) to remove redundancy.
% From the real defocus blur in CataBlur, the simulated defocus blur in CaDISBlur is proven to be highly realistic.
% The CataBlur dataset itself also demonstrates the severity of defocus blur in microscope surgery that poses strong degradation to the image, which leads to serious performance drops in downstream tasks, as shown in \cref{sec:experiments}.

% TODO gaussian kernel size, why CaDISBlur?

\begin{table}[t]
  \centering
  \setlength{\tabcolsep}{0.7mm}{} 
  \scalebox{0.93}{
  \begin{tabular}{lcccc}
    \toprule
Scenario & Dataset & \#Image & Resolution & Label\\
    \midrule
\multirow{3}{*}{\makecell[l]{Real\\ world}}&DPDD \cite{abuolaim_defocus_2020} & 500& 1680$\times$1120& Labeled\\
                        &LFDOF \cite{ruan_aifnet_2021} & 12,826& 1008$\times$688 & Labeled\\
                        &CUHK \cite{shi_discriminative_2014} & 704& $\sim$ 640$\times$480& Unlabeled\\
    \midrule
\multirow{3}{*}{\makecell[l]{Cell\\ microscopy}} &BBBC006 \cite{ljosa_annotated_2012} & 6144& 696$\times$520& Labeled\\
                                &3DHistech \cite{geng_cervical_2022} & 94,973& 256$\times$256& Labeled\\
                                &WNLO \cite{geng_cervical_2022} & 108,065& 256$\times$256& Unlabeled\\
    \midrule
\multirow{2}{*}{\makecell[l]{Surgical\\ microscopy}} &CaDISBlur& 9340& 960$\times$540& Labeled\\
                                 &CataBlur & 1208& 1280$\times$720& Unlabeled\\
    \bottomrule
  \end{tabular}
  }
  \caption{Dataset description.}
  \label{tab:dataset}
\end{table}

\section{Supplementary Experiments}
\label{sec:sup_experiment}
% \paragraph{CaDISBlur}
% TODO result on blur instrument and blur anatomies, respectively?

\paragraph{Data deficiency}
As addressed in \cref{sec:intro}, the data deficiency in microscopy datasets can pose harm to generalizability.
% The experiments in \cref{sec:experiments} prove that learning rich information from real-world dataset can enhance microscopy deblur.
To further demonstrate the drawbacks brought by data deficiency, experiments are conducted on unlabeled microscopy datasets WNLO \cite{geng_cervical_2022} and CataBlur regarding two settings. 
For the first setting $S_1$, the MPT is \textbf{trained on intra-domain microscopy images and tested on unlabeled microscopy datasets}, i.e., trained on 3DHistech \cite{geng_cervical_2022} then tested on WNLO \cite{geng_cervical_2022}, and trained on CaDISBlur then tested on CataBlur.
For $S_2$, the MPT is \textbf{trained on cross-domain real-world images (LFDOF \cite{ruan_aifnet_2021}) and tested on unlabeled microscopy datasets}.

\begin{figure*}[!h]
  \centering
   \includegraphics[width=1\linewidth]{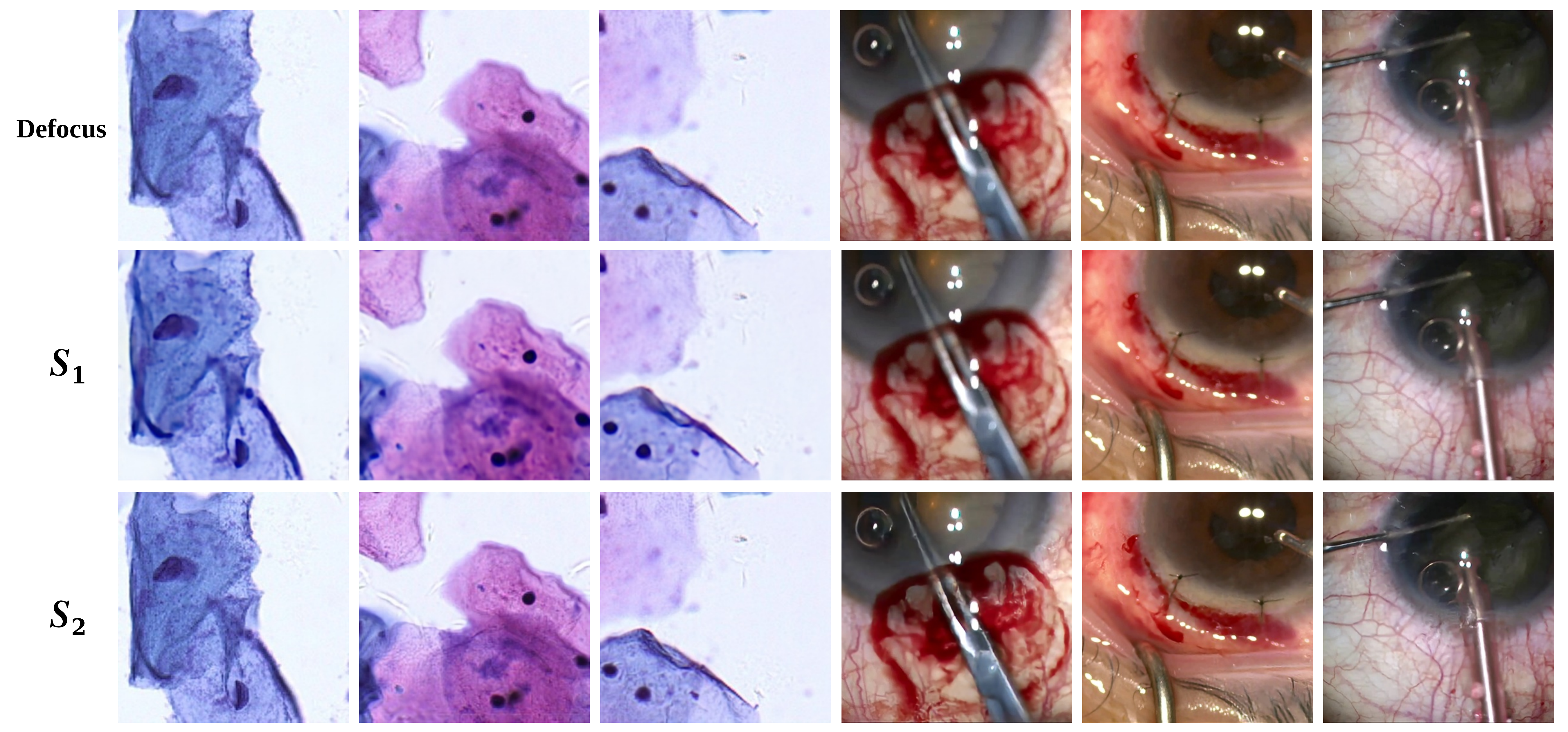}
   \caption{Illustration of models trained with data deficient microscopy dataset ($S_1$) and data sufficient real-world dataset ($S_2$), including unlabeled dataset WNLO \cite{geng_cervical_2022} (left) and CataBlur (right). Even if the model is trained with intra-domain microscopy data ($S_1$), the deblur restoration turns out to be trivial for CataBlur, or even brings strong artifacts for WNLO. As for $S_2$, the deblur result tends to have fewer artifacts, and more fine details are restored. It proves the existence of data deficiency problem and the necessity of learning cross-domain knowledge. \vspace{-5mm}}
   \label{fig:sup_FD}
\end{figure*}

The deblur results are shown in \cref{fig:sup_FD}, from which it can be observed that the model trained with cross-domain real-world dataset ($S_2$) leads to fewer artifacts and more fine details in its deblurred results than the model trained with intra-domain microscopy dataset ($S_1$). 
This phenomenon proves the existence of data deficiency in microscopy dataset, i.e., the model trained with microscopy dataset suffers from poor generalizability caused by the insufficient features contained in microscopy dataset.
From visualization of $S_2$, the necessity of learning cross-domain rich deblur guidance is also proved.

\paragraph{Ablation studies on CSWA and FEFN}

\begin{figure}[!h]
  \centering
   \includegraphics[width=0.9\linewidth]{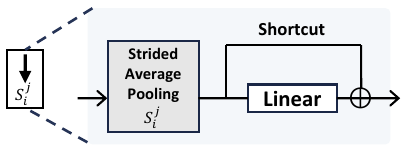}
   \caption{Downsampling module adopted in CSWA. \vspace{-3mm}}
   \label{fig:sup_downsamp}
\end{figure}

\begin{figure}[!h]
  \centering
   \includegraphics[width=1\linewidth]{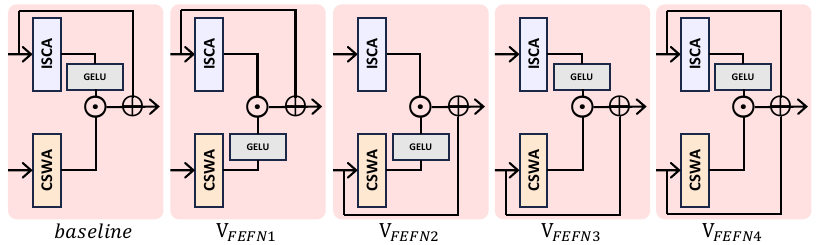}
   \caption{Baseline and variant design of ablation studies for FEFN. V$_{FEFN1}$ is the reversed structure (V$_3$) shown in \cref{tab:ablationfefn} \vspace{-3mm}}
   \label{fig:sup_fefn}
\end{figure}

\begin{table}[!h]
  \centering
  \setlength{\tabcolsep}{0.8mm}{} 
  \scalebox{1}{
  \begin{tabular}{lccc}
    \toprule
 \multicolumn{1}{l}{Configuration}  & \multicolumn{1}{c}{$\Delta$PSNR$_D$} & $\Delta$PSNR$_B$& $\Delta$PSNR$_C$ \\
    \midrule
 $V_{ds1}$ (w/o shortcut)& -0.14& -0.02&-0.15\\
 $V_{ds2}$ (w/o linear)& -0.05& -0.04&-0.07\\
 \multicolumn{1}{l}{$V_{ds3}$ (max pooling)}  & \multicolumn{1}{c}{-0.37} & -0.49&  -0.63\\
 \multicolumn{1}{l}{$V_{ds4}$ (convolution)}  & \multicolumn{1}{c}{-0.03} & -0.09&  -0.06\\
 $V_{ds5}$ (interpolate)& -0.01& +0.02&-0.04\\
 $V_{w/o~NPC}$& -0.09& -0.07&-0.12\\
 $V_{w/o~sw}$& -0.06& -0.13&-0.09\\
  % $V_{QKV}$ (reversed $Q$-$KV$)& & -0.25&  \\
      \midrule
   $V_{FEFN1}$& -0.23& -0.17&-0.48\\
   $V_{FEFN2}$& -0.09& -0.12&-0.15\\
   $V_{FEFN3}$& -0.60& -0.74&-0.91\\
   $V_{FEFN4}$& -0.24& -0.31&-0.20\\
    \bottomrule
  \end{tabular}
  }
  \caption{Ablation studies on CSWA and FEFN. $\Delta$PSNR refers to the change in PSNR compared with the MPT baseline.}
  \label{tab:sup_ablationCSWAFEFN}
\end{table}
% \vspace{-6mm}
% Ablation of DownSamp module, NPConv, shift window, Q/KV reversed, shortcut in FEFN

Additional ablation studies regarding the structure of the proposed CSWA and FEFN are conducted.
The results are shown in \cref{tab:sup_ablationCSWAFEFN}.
For the downsampling module adopted in CSWA (\cref{fig:sup_downsamp}), a series of variants are evaluated, including variant without shortcut connection ($V_{ds1}$), the variant without linear projection and shortcut connection ($V_{ds2}$), and variants changing the downsampling operation from strided average pooling to strided maximum pooling ($V_{ds3}$), strided convolution ($V_{ds4}$), and bicubic interpolation ($V_{ds5}$).
The results in \cref{tab:sup_ablationCSWAFEFN} show that the current baseline outperforms the variants in most of the situations.
Although $V_{ds5}$, which adopts bicubic interpolation, achieves almost the same performance as the baseline or even trivial improvement, it leads to complex computation that hinders parallel training and inference.  
Ablation studies on NPConv and shifting window mechanism in CSWA are also carried out, which are denoted by $V_{w/o~NPC}$ and $V_{w/o~sw}$, respectively.
The results demonstrate the superiority of our design.

Going further from the experiments in \cref{sec:discussion}, more ablation studies regarding the asymmetric activation mechanism and shortcut connection in FEFN are conducted, including four variants as shown in \cref{fig:sup_fefn}.
Based on the result reported in \cref{tab:sup_ablationCSWAFEFN}, it can be concluded that the baseline structure adopted in this paper achieves the best performance among all the variants.

\paragraph{Ablation studies on pyramid scales}
Following the description in \cref{sec:experiments4.1} to keep the sub-block number, feature dimensions, and attention heads unchanged, ablation studies on pyramid scales ($S_i$) are carried out regarding variants with  similar FLOPs and parameters:

    1) V$_1$: $S_1=S_2=S_3=S_4=[1,1,1,1,1,1]$. In this variant, CSWA actually downgrades to the original WA (the variant with WA+ISCA shown in \cref{tab:ablationcswa}), and the image pyramid is not constructed. 
    
    2) V$_2$: $S_1=[\frac{1}{16},\frac{1}{16},\frac{1}{8},\frac{1}{8},1,1]$, $S_2=[\frac{1}{8},\frac{1}{8},\frac{1}{4},\frac{1}{4},1,1]$, $S_3=S_4=[\frac{1}{4},\frac{1}{4},\frac{1}{2},\frac{1}{2},1,1]$. This variant explores the pyramid structure with smaller scales than the baseline.

    3) V$_3$: $S_1=[\frac{1}{4},\frac{1}{4},\frac{1}{2},\frac{1}{2},1,1]$, $S_2=[\frac{1}{2},\frac{1}{2},\frac{1}{2},\frac{1}{2},1,1]$, $S_3=S_4=[\frac{1}{2},\frac{1}{2},\frac{1}{2},\frac{1}{2},1,1]$. This variant adopts larger-scale pyramids than the baseline.

As shown by the results in \cref{tab:sup_abblationpyramid}, all the variants lead to performance drops. For V$_1$ with no pyramid structure, significant performance degradation is observed on BBBC006, which is the dataset with one of the longest attention spans. 
A similar phenomenon is observed in SwinIR \cite{liang_swinir_2021} that does not feature a multi-scale pyramid.
It achieves inferior performance on BBBC006 as shown in \cref{tab:resultmicro}.
These together prove the effectiveness of our multi-pyramid design, especially for the microscopy deblur tasks.
The performance degradation in V$_2$ and V$_3$ shows the superiority of the pyramid scales in the baseline model.
\begin{table}[!h]
  \centering
  \setlength{\tabcolsep}{0.8mm}{} 
  \begin{tabular}{cccc}
    \toprule
 \multicolumn{1}{c}{Pyramid}  & \multicolumn{1}{c}{$\Delta$PSNR$_D$} & $\Delta$PSNR$_B$& $\Delta$PSNR$_C$\\
    \midrule
   V$_1$&  -0.08& -0.79&  -0.20\\
   V$_2$&  -0.05& -0.02&  -0.07\\
  V$_3$& -0.01& -0.04&  -0.03\\
    \bottomrule
  \end{tabular}
  \caption{Ablation studies on pyramid scales.}
  \label{tab:sup_abblationpyramid}
\end{table}

% \paragraph{TODO}
% TODO: Ablation pretrain and fine tuning with L1loss and VGGloss,  pretrian/fune-tune or train together

% TODO: Ablation unlabeled image deblur, EFCR, use different data for training
% Ablation: relative extended CR (normalization term)

% TODO Average attention distance

% compared with vanilla dataset (uniformly blurring the whole image), our synthesized enforce the model learning edge information, which is critical for segmentation

\begin{figure*}[!h]
  \centering
   \includegraphics[width=1\linewidth]{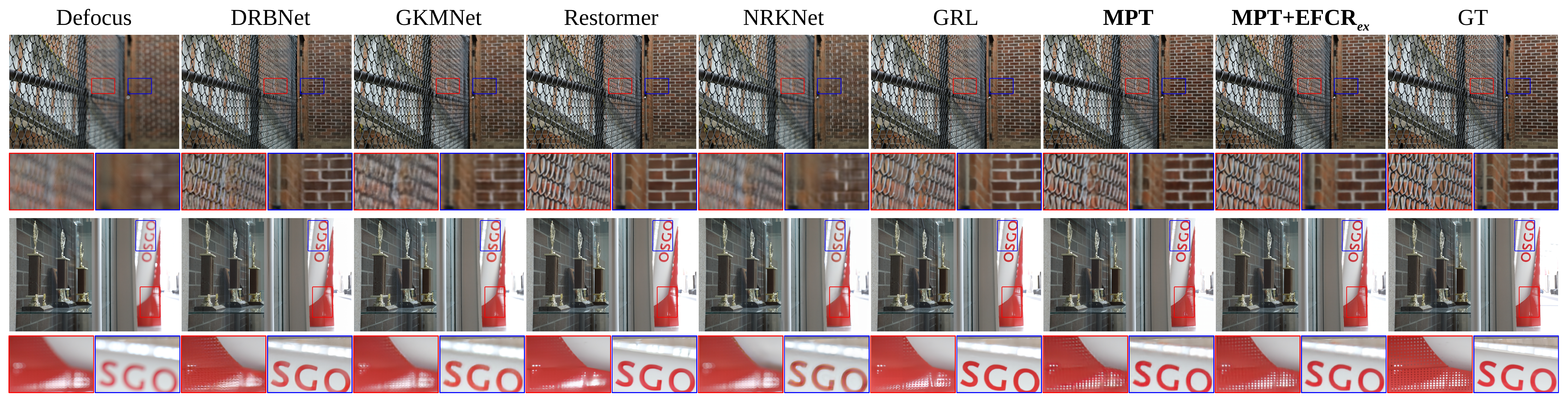}
   \caption{Visualization of deblur on DPDD \cite{abuolaim_defocus_2020} dataset. As the comparison shows, our MPT achieves the best deblur performance in real-world scenes, restoring most of the small-scale fine details and large-scale patterns. With the help of EFCR$_{ex}$, the performance is further enhanced. It shows the superiority and generalizability of our framework.}
   \label{fig:sup_DPDD}
\end{figure*}

\begin{figure*}[!h]
  \centering
  \begin{subfigure}{1\linewidth}
    \includegraphics[width=0.95\linewidth]{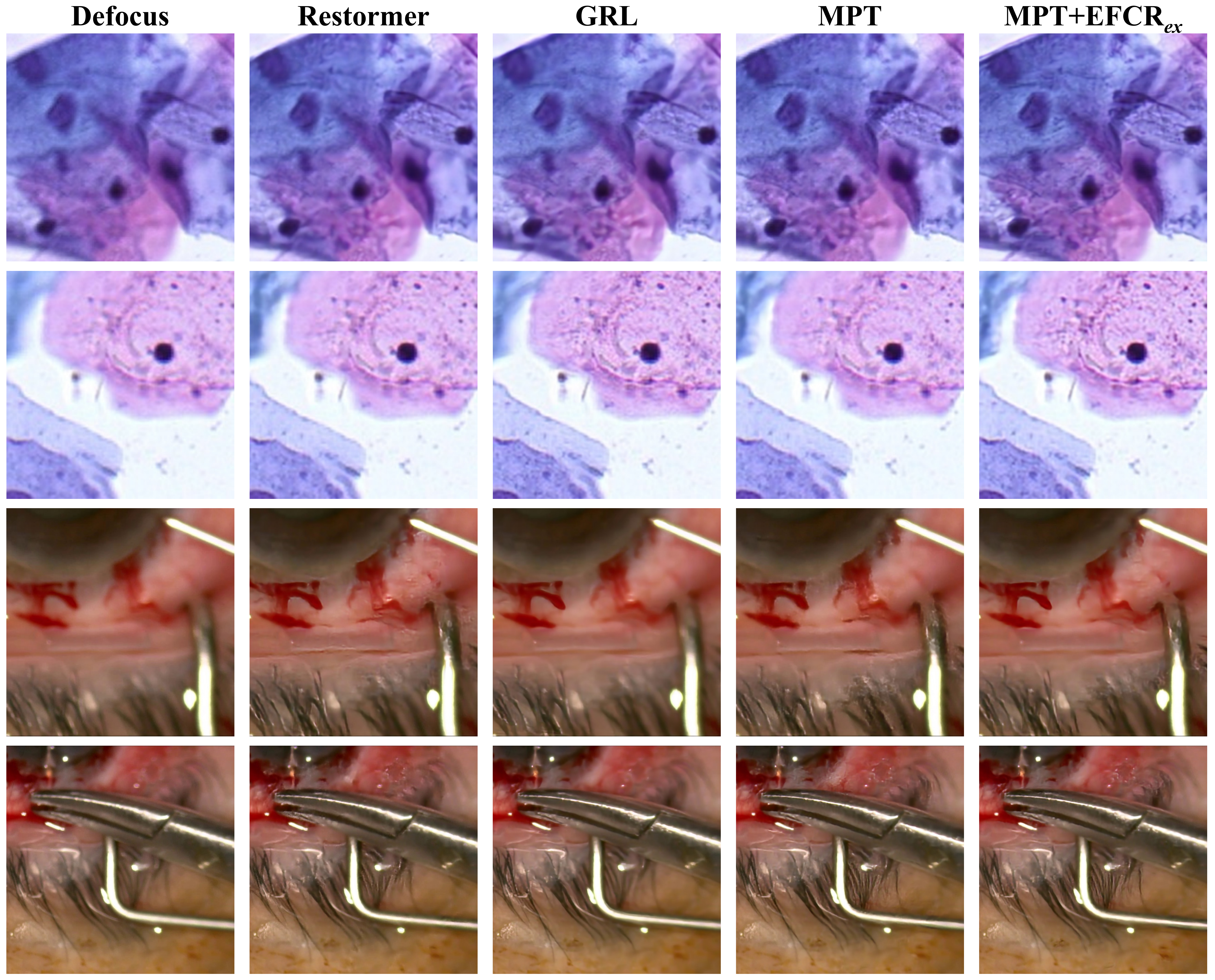}
    \caption{Visualization of deblur on WNLO \cite{geng_cervical_2022} (top) and CataBlur (bottom) datasets.}
    \label{fig:sup_wnlocatablur}
  \end{subfigure}
  
  \begin{subfigure}{1\linewidth}
    \includegraphics[width=0.95\linewidth]{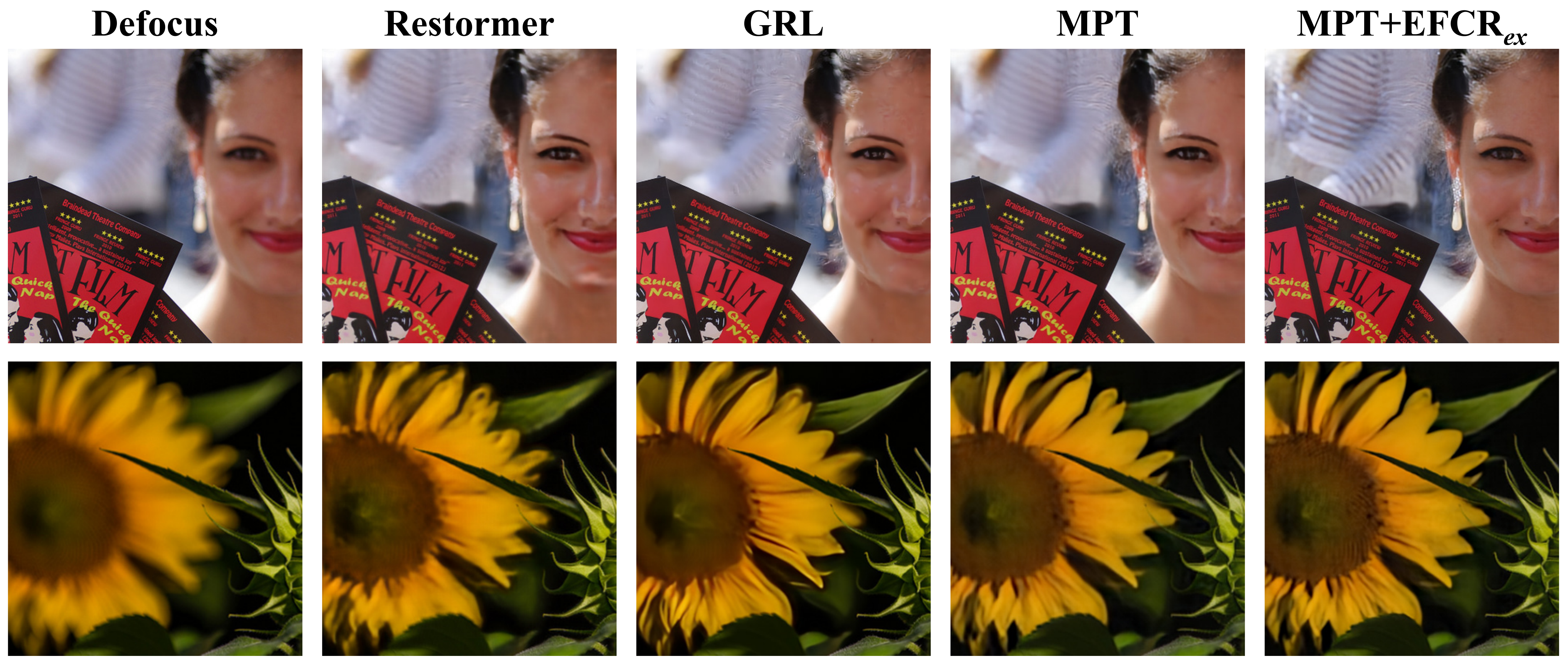}
    \caption{Visualization of deblur on CUHK \cite{shi_discriminative_2014} dataset.}
    \label{fig:sup_CUHK}
  \end{subfigure}
  \caption{Visualization of unsupervised deblur on CataBlur, WNLO \cite{geng_cervical_2022}, and CUHK \cite{shi_discriminative_2014}. The proposed framework achieves the best unsupervised deblur performance on both microscopy datasets and real-world datasets, showing high generalizability. By further applying EFCR$_{ex}$, the deblur performance is significantly enhanced, proving the effectiveness of the proposed EFCR for knowledge transfer.}
\end{figure*}

\paragraph{Additional visualization}
Demonstrations of supervised real-world deblur are shown in \cref{fig:sup_DPDD} for labeled datasets DPDD \cite{abuolaim_defocus_2020}.
Illustrations of unsupervised deblur on microscopy dataset and real-world dataset are provided in \cref{fig:sup_wnlocatablur} and \cref{fig:sup_CUHK}, respectively.
The visualization proves that our method achieves the best performance on microscopy datasets and real-world datasets regarding various patterns in both supervised and unsupervised scenarios.
More visualizations of the results of cell detection on BBBC006 \cite{ljosa_annotated_2012} and surgical scene semantic segmentation on CaDISBlur are provided in \cref{fig:sup_seg}, from which it can be concluded that the downstream tasks results on deblurred images from our method achieves more satisfactory outcomes.

% TODO: additional eg for WNLO and CataBlur

% \section{Conclusion and Limitations}
% \label{sec:sup_limit}
% The experiment results and visualization shown in the supplementary material validate the effectiveness and superiority of the proposed framework on not only microscopy tasks but also real-world tasks.

\begin{figure*}[!h]
  \centering
  \begin{subfigure}{1\linewidth}
    \includegraphics[width=1.0\linewidth]{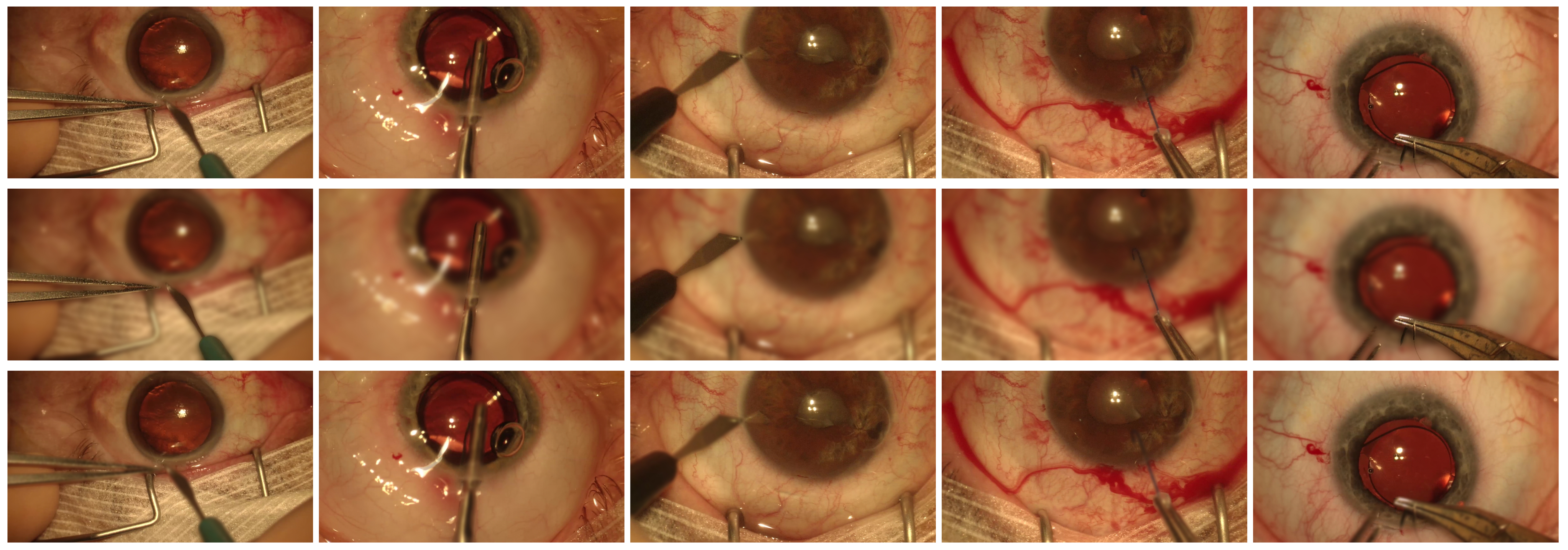}
    \caption{Examples for CaDISBlur dataset, where the top row is the original sharp images (\textbf{ground truth}), the middle row and bottom row are the corresponding blurry images with the focal plane on instruments (\textbf{blurry anatomies}) or anatomies (\textbf{blurry instruments}), respectively.}
    \label{fig:sup_cataexamples-a}
  \end{subfigure}
  
  \begin{subfigure}{1\linewidth}
    \includegraphics[width=1.0\linewidth]{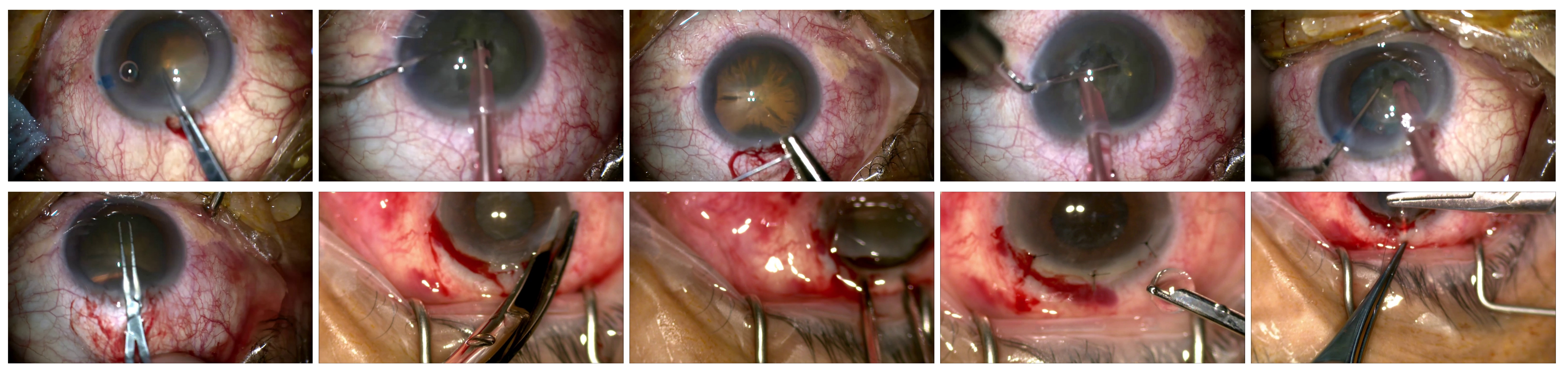}
    \caption{Examples for CataBlur dataset, which is an unlabeled dataset containing only defocus images.}
    \label{fig:sup_cataexamples-b}
  \end{subfigure}
  \caption{Examples of samples in CaDISBlur and CataBlur.}
  \label{fig:sup_cataexamples}
\end{figure*}

% \begin{figure*}[!h]
%   \centering
%    \includegraphics[width=1\linewidth]{fig/sup_CUHK.pdf}
%    \caption{Visualization of deblur on CUHK \cite{shi_discriminative_2014} dataset. The proposed framework achieves the best unsupervised deblur performance, showing the extraordinary generalizability. By further applying EFCR$_{ex}$, the deblur performance is significantly enhanced, proving the effectiveness of the proposed EFCR for knowledge tansfer.}
%    \label{fig:sup_CUHK}
% \end{figure*}

\definecolor{pupilcolor}{HTML}{0089FF}
\definecolor{iriscolor}{HTML}{FF0000}
\definecolor{corneacolor}{HTML}{FFFFFF}
\definecolor{skincolor}{HTML}{FF00A5}
\definecolor{tapecolor}{HTML}{FFA500}
\definecolor{retractorcolor}{HTML}{6300FF}
\definecolor{instcolor}{HTML}{8D8D8D}
\definecolor{handcolor}{HTML}{FF9CC9}

\begin{figure*}[!h]
  \centering
   \includegraphics[width=1\linewidth]{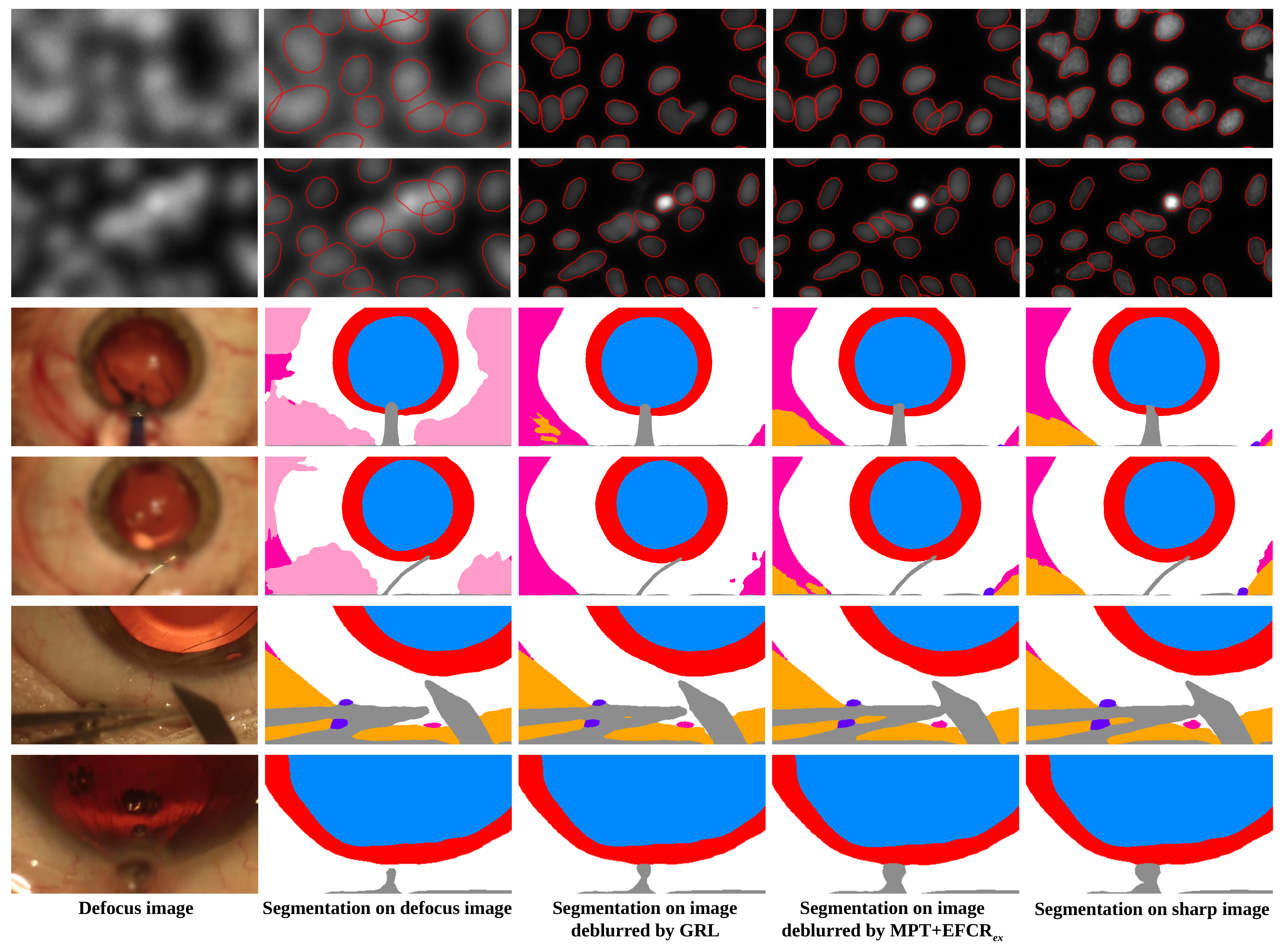}
   \caption{Results of downstream tasks on BBBC006 \cite{ljosa_annotated_2012} (top) and CaDISBlur (bottom). For BBBC006, our method can restore precise cell shape with sharper outline, achieving more accurate segmentation and detection. For CaDISBlur, the deblurred images by our framework contain more differentiable features, leading to more accurate semantic segmentation in both anatomies and instruments.
   (Colormap: \sethlcolor{pupilcolor}\hl{\qquad\qquad} Pupil, \sethlcolor{iriscolor}\hl{\qquad\qquad} Iris, \sethlcolor{handcolor}\hl{\qquad\qquad} surgeon's hand, \fbox{\phantom{ccc}} Cornea, \sethlcolor{skincolor}\hl{\qquad\qquad} Skin, \sethlcolor{tapecolor}\hl{\qquad\qquad} Surgical tape, \sethlcolor{retractorcolor}\hl{\qquad\qquad} Eye retractors, \sethlcolor{instcolor}\hl{\qquad\qquad} Instruments)}
   \label{fig:sup_seg}
\end{figure*}

\end{document}